\documentclass{article}

\usepackage[nonatbib,preprint]{neurips_2026}
\usepackage[square,numbers,sort&compress]{natbib}
\usepackage{bibunits} 

\usepackage{enumitem}
\usepackage[section]{placeins}
\usepackage{float}

\usepackage[utf8]{inputenc} 
\usepackage[T1]{fontenc}    
\usepackage{hyperref}       
\usepackage{url}            
\usepackage{booktabs}       
\usepackage{amsfonts}       
\usepackage{nicefrac}       
\usepackage{microtype}      
\usepackage{xcolor}         
\usepackage{graphicx}       
\usepackage{subcaption}

\usepackage{amsmath}
\usepackage{amssymb}
\usepackage{mathtools}
\usepackage{amsthm}

\usepackage[nonumberlist]{glossaries}
\makenoidxglossaries

\newglossaryentry{Voxel}
{
    name=Voxel,
    description={Volumetric pixel. The basic neural unit in fMRI; voxels are typically  1-2 mm\textsuperscript{3} isometric cubes that tile the brain uniformly. A voxel may contain several hundred thousand neurons \cite{huettelFunctionalMagneticResonance2004}}
}
\newglossaryentry{TR}
{
    name=TR,
    description={Repetition Time. The sampling rate in fMRI, typically around 0.5-1 Hz. That is, a single 3-D volume of whole-brain functional activity is collected every 1-2 seconds}
}
 \newglossaryentry{BOLD}
{
    name=BOLD,
    description={Blood Oxygen Level-Dependent response. Rather than directly measuring electrical neural activity, fMRI measures changes in blood flow (primarily via the movement of deoxygenated hemoglobin) as an indirect proxy for neural activity. The temporally-extended BOLD response occurs on the order of seconds, peaking approximately 6-12 seconds after the underlying neural spikes \cite{huettelFunctionalMagneticResonance2004}}
}
\newglossaryentry{HRF}
{
    name=HRF,
    description={Hemodynamic Response Function. A function used to model the time course of the BOLD response that is triggered by neural activity. The HRF may vary as a function of subject, voxel, and stimulus, among other factors}
}
\newglossaryentry{GLM}
{
    name=GLM,
    description={General Linear Model. A common step in fMRI data analysis that estimates the response strength of a particular brain voxel or region of interest to a single stimulus. The temporally-extended BOLD response unfolds over seconds; the GLM models the stimulus response using an HRF so that we can account for this lag and convert a timeseries of measurements into a single response estimate per stimulus \cite{andrew_jahn_2022_5879294}}
}
\newglossaryentry{Betas}
{
    name=Single-trial response estimates/betas,
    description={The regression beta coefficients per voxel obtained from the GLM. The GLM yields a single vector per stimulus, where the value of each element represents our estimate of each voxel or brain region's response to one stimulus. These betas (as opposed to the fMRI timeseries itself) are the inputs to the MindEye2 architecture}
}

\newglossaryentry{Closed-loop Neurofeedback}
{
    name=Closed-loop Neurofeedback,
    description={A neuroimaging paradigm in which stimulus presentation is modulated based on a real-time readout of the participant's brain activity, with the goal of driving brain activity towards a particular state \cite{sitaramClosedloopBrainTraining2017,lubianikerNeurofeedbackLensReinforcement2022a}. This often involves providing a visual signal (such as a thermometer) that indicates the extent to which the participant's brain activity resembles the target state}
}

\usepackage[disable,textsize=tiny]{todonotes}

\title{Real-time Reconstruction of Human Visual Perception from fMRI}

\author{%
  Rishab S. Iyer\textsuperscript{1} \\
  Princeton University \\
  \texttt{rsiyer@princeton.edu} \\
  \AND
  Jiaxin Cindy Tu\textsuperscript{*, 1} \\
  Dartmouth College \\
  \And
  Cesar Kadir Torrico Villanueva\textsuperscript{*, 1} \\
  \And
  Anish Mahishi\textsuperscript{*, 1} \\
  \AND
  Ross P. Kempner\textsuperscript{1} \\
  Icahn School of Medicine at Mount Sinai \\
  \And
  Jacob S. Prince \\
  Harvard University \\
  \AND
  Ernest W. Lo\textsuperscript{1} \\
  \And
  Akash Bhowmick\textsuperscript{1} \\
  Princeton University \\
  \And
  Hritik Arasu\textsuperscript{1} \\
  University of Texas at Dallas \\
  \And
  Amaar Chughtai\textsuperscript{1} \\
  \AND
  \begin{minipage}[t]{0.30\textwidth}\centering\bfseries
  Elizabeth A. McDevitt \\
  \mdseries Princeton University \\
  \end{minipage}\hspace{0.5em}%
  \begin{minipage}[t]{0.30\textwidth}\centering\bfseries
  Paul S. Scotti\textsuperscript{\dag, 1} \\
  \mdseries \href{https://sophont.med}{Sophont} \\
  \mdseries Princeton University \\
  \end{minipage}\hspace{0.5em}%
  \begin{minipage}[t]{0.30\textwidth}\centering\bfseries
  Kenneth A. Norman\textsuperscript{\dag} \\
  \mdseries Princeton University \\
  \end{minipage}
  \AND
  \begin{minipage}{0.95\textwidth}\centering\normalfont
  \textsuperscript{1}\href{https://www.medarc.ai/}{Medical AI Research Center (MedARC)} \quad
  \textsuperscript{*}Core contribution \quad
  \textsuperscript{\dag}Co-senior authors \quad
  \end{minipage}
}

\begin{document}

\maketitle

\begin{abstract}
  Real-time closed-loop neurofeedback based on functional magnetic resonance imaging (fMRI) has led to important scientific and clinical advances. However, the sophistication of the analysis methods used in real-time fMRI lags behind the state-of-the-art in fMRI decoding, largely due to computational factors: Most advanced decoding pipelines do not fit within the envelope of real-time processing, where the analysis needs to be conducted in a matter of seconds and without leveraging data acquired later in the session. Here, we present a real-time compatible adaptation of a computationally intensive state-of-the-art pipeline for reconstructing perceived natural images (MindEye2), and we demonstrate that reliable fine-grained decoding is still achievable in this setting. Using RT-Cloud, an open-source, scalable cloud-based platform, we performed a real-time scan where we decoded single-trial visual perception within seconds after an image was shown to the participant. Finally, we use simulated analyses to document the factors driving changes in performance from offline to real-time analysis. This work serves as a proof-of-concept that it is feasible to deploy these powerful fMRI decoding pipelines in real-time analysis, paving the way for their use in brain-computer interfaces for scientific discovery and clinical treatment.
\end{abstract}

\begin{bibunit}

\section{Introduction}
A longstanding goal for cognitive neuroscience is to develop non-invasive methods for decoding thoughts based on brain activity -- the finer-grained the decoding, the more useful this information is likely to be for downstream scientific and clinical applications. Of all the non-invasive brain imaging methods, functional magnetic resonance imaging (fMRI) holds the greatest promise of enabling such fine-grained decoding of cognitive states, given its superior spatial resolution compared to other non-invasive methods like electroencephalography (EEG) and magnetoencephalography (MEG). Indeed, in recent years there have been rapid advances in our ability to decode fine-grained sensory information from fMRI, largely driven by the advent of foundation models in computer vision \cite{radfordLearningTransferableVisual2021} and natural language processing \cite{radfordLanguageModelsAre2019,devlinBERTPretrainingDeep2019,touvronLLaMAOpenEfficient2023}. These advances in representation learning have resulted in impressive feats of mind-reading: reconstructing the images that a participant has seen \cite{takagiHighresolutionImageReconstruction2023a,ferranteTheirEyesMultisubject2024,ozcelikReconstructionPerceivedImages2022,ozcelikNaturalSceneReconstruction2023a,scottiReconstructingMindsEye2023a,scottiMindEye2SharedSubjectModels2024,gongMindTunerCrossSubjectVisual2024} or imagined \cite{kneelandNSDImageryBenchmarkDataset2025}, or the contents of a narrative that they listened to \cite{tangSemanticReconstructionContinuous2023}. 

\begin{figure*}[t!]
  \centering
  \includegraphics[keepaspectratio, width=\textwidth]{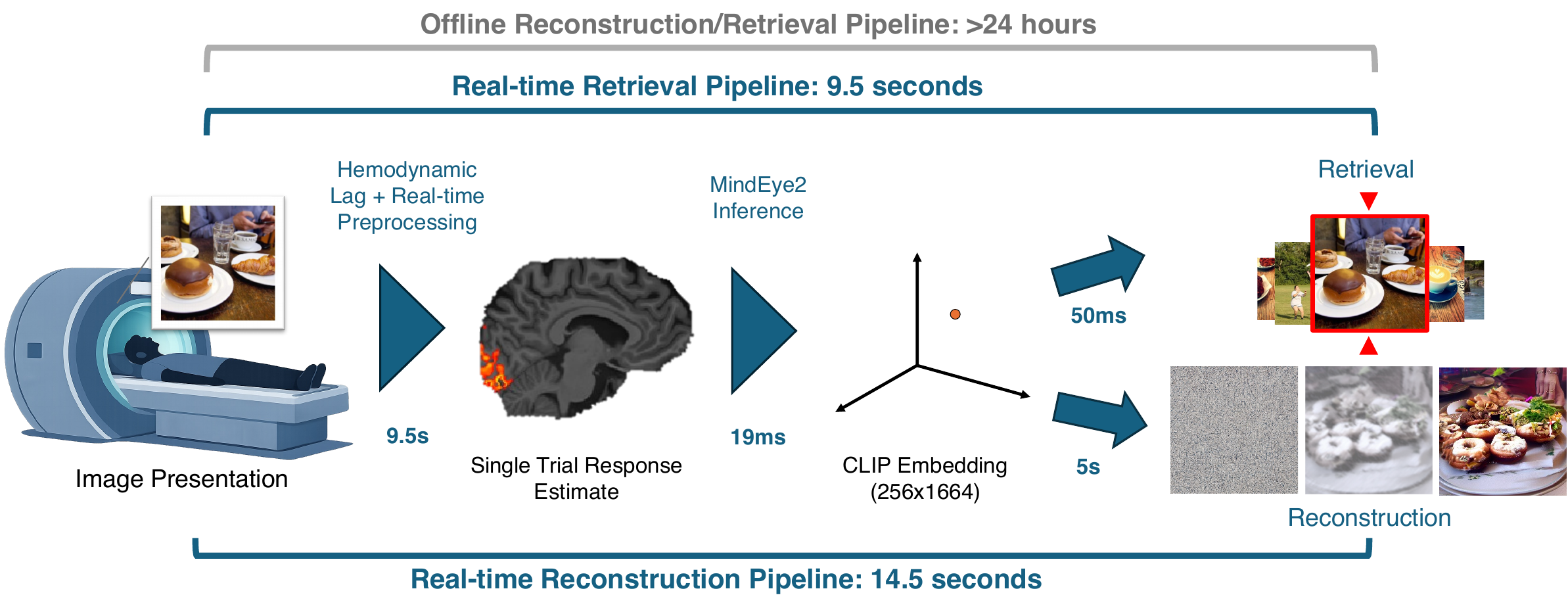}
  \caption{Real-time fMRI-to-image decoding pipeline. The participant views an image in the scanner; the BOLD signal undergoes real-time compatible preprocessing and a GLM extracts a single-trial response estimate (beta). MindEye2 maps betas to CLIP embeddings, which can then be used to reconstruct the seen image or retrieve the image out of a pool of candidates within seconds after it is viewed.}
  \label{fig:pipeline}
\end{figure*}

In parallel, there have also been advances in deploying fMRI in real-time settings, such as in \glslink{Closed-loop Neurofeedback}{closed-loop neurofeedback} paradigms where participants are guided towards targeted brain states by feedback based on \glslink{BOLD}{BOLD} (blood oxygen level-dependent) signals that are processed on-the-fly \cite{sitaramClosedloopBrainTraining2017,motiwalaBrainComputerInterfaces2026}. These studies have provided important insights into clinical conditions such as depression \cite{youngRandomizedClinicalTrial2017,mennenCloudBasedFunctionalMagnetic2021a} and have also shed light on fundamental learning mechanisms in the brain \cite{pengInducingRepresentationalChange2024}. However, in comparing recent real-time fMRI studies to state-of-the-art methods used for decoding in offline analyses, there is a clear disconnect. The multivariate analysis methods used in recent real-time studies (e.g., decoding whether people are attending to a scene or a face) are sophisticated compared to prior methods in fMRI neurofeedback (e.g., tracking the average activation of the amygdala), but they are nowhere near as sophisticated as state-of-the-art fMRI decoding methods that enable fine-grained reconstruction of individual stimuli. Importing such sophisticated decoding methods into real-time fMRI pipelines would vastly improve their utility as brain-computer interfaces, enabling novel neurofeedback experimental designs (e.g., allowing patients with depression to see how their perception of an image overemphasizes negative features of the image). The main obstacle to achieving this goal is computational. State-of-the-art fMRI decoding methods are computationally intensive, often requiring extensive data processing that takes hours to days. Additionally, decoders need to be adapted to generalize to the inherent structural constraints of real-time fMRI: conducting the analysis within seconds, without access to the data collected later in the session. 

In the present work, we chose reconstruction of seen images as the ideal use-case to demonstrate the feasibility of adapting a computationally heavy machine learning analysis for real-time fMRI. The pipeline reported here is (to our knowledge) the first to achieve real-time reconstruction of visual perception from fMRI. To accomplish this, we use RT-Cloud \cite{wallaceRTCloudCloudbasedSoftware2022a}, a scalable open-source framework that enables real-time streaming and analysis of fMRI data, in conjunction with MindEye2 \cite{scottiMindEye2SharedSubjectModels2024}, a large (\textgreater{}700M parameters) machine learning architecture for visual decoding. We conducted tests with a 3 Tesla (3T) MRI scanner, a platform available at tens of thousands of facilities at hospitals and universities worldwide, rather than the less prevalent 7 Tesla (7T) platform that supported much prior work in this space. We demonstrate that, by fine-tuning our model with just one hour of data from a new participant, our pipeline can support fine-grained visual decoding in a subsequent session, reconstructing images within seconds after they are viewed by the participant. 

After developing these methods in simulation using previously collected 3T data, we successfully implemented this in a real-time fMRI session (infrastructure details in Appendix \ref{rt-infra}) where we displayed reconstructions soon after the participant viewed natural images. Here, we present simulated real-time analyses to test variations in preprocessing and analysis pipelines. 

The main contributions of this work are as follows:
\begin{itemize}[leftmargin=*]
    \item  We demonstrate for the first time that it is possible to decode seen images from fMRI at single-trial resolution in real-time (as fast as 9.5 seconds post image onset for retrieval), improving on prior approaches that span hours to days. 
    \item We show that reliable real-time decoding is achievable with minimal training data ($\sim$1 hour) from a new participant and can be generalized from 7T to 3T fMRI.
    \item We compare several real-time setups that trade off decoding performance with latency, quantifying the changes in performance one can expect transitioning from offline to real-time analysis.
    \item We make all code, models, and data publicly available\footnote{\url{https://github.com/brainiak/rtcloud-projects/tree/main/mindeye}} and carry out development fully in-the-open to encourage collaboration and downstream applications. 
\end{itemize}

\section{Methods}
In this section, we describe step-wise modifications to our pipeline, starting with offline analysis \cite{scottiMindEye2SharedSubjectModels2024} and working our way down to the fastest real-time analysis (fMRI has intrinsic temporal limitations; see Section \ref{rtpreproc}). This serves to document the progressive trade-off between speed and performance. 

Training and evaluation are kept constant across pipeline variations; this facilitates fair comparisons and enables us to attribute differences in performance to the components of the pipelines. To this end, all of the results shown below involve pretraining on data from 7 participants from the Natural Scenes Dataset (NSD; \citealp{allenMassive7TFMRI2022}); we fine-tune on only one session of 3T fMRI data ($\sim$1 hour) from a new participant, and we evaluate on a subsequent session from the same participant. Additionally, we replicate key findings using the held-out NSD subject. Overall, these results highlight how we are able to achieve above-chance real-time decoding for a new participant as early as their second scanning session. 

\subsection{3T fMRI Data Collection}
\subsubsection{Participant}
The 3T fMRI data were collected from an author of this study (R.S.I., age 23, male). The subject was right-handed, healthy, and had normal vision. Written informed consent was obtained from the participant and the experimental protocol was approved by Princeton University's Institutional Review Board. The participant was compensated at a standard rate of \$30 per hour, plus bonuses for return visits to the scanner.

\subsubsection{Stimuli}
The stimuli used in this study were natural images (such as locations, animals, and people); most of these images were from NSD, which itself sampled from Microsoft COCO \cite{linMicrosoftCOCOCommon2015}. 

Both scan sessions (training and real-time) involved viewing 531 unique images. This included 50 images from the “special515” subset of NSD images, each presented three times; 419 additional images from NSD, each presented once; and 31 pairs of highly similar images (e.g., two lighthouses), where each of these 62 items was presented twice. The paired images (some from \citealp{wanjiaAbruptHippocampalRemapping2021} and some generated using StyleGAN; \citealp{karrasStyleBasedGeneratorArchitecture2019}) were included to support future investigations of the model's ability to distinguish fine-grained representations but were not analyzed further here (although they were included in the training data for the model). Including repetitions, this totaled 693 image presentations per scan. Only the 31 pairs of similar images (repeated twice each) were presented in both sessions; all other images differed between the two sessions. 

\subsubsection{Experimental Design}
We employed a rapid event-related design closely matched to NSD. Each session consisted of 11 functional runs of 63 images presented every 4 seconds. Each image appeared for 3 seconds with a 1-second inter-stimulus interval. The order of presentation was pseudo-random with a constraint to prevent back-to-back repeated presentations of the same image. Blank trials were interspersed to assist with response estimation (see Section \ref{discussion-offline-vs-realtime}). 

The subject was instructed to fixate on a small red dot that was present throughout the experiment, and to avoid eye movements even if drawn to look at details in different parts of an image. When an image was presented, the subject's task was to determine whether the image was ``new'' or ``old'', responding with a press of ``1'' or ``2'' on a button-box, respectively. At the end of each run, they received accuracy feedback. This familiarity detection task served as a way to maintain the subject's attention (as in  \citealp{allenMassive7TFMRI2022}) and responses were not analyzed. 


\subsection{3T fMRI Acquisition}
MRI data were collected with a 3T Siemens Prisma scanner with a 64-channel head coil located at the Scully Center for the Neuroscience of Mind and Behavior at the Princeton Neuroscience Institute. Functional scans were acquired using a T2*-weighted multiband EPI sequence (repetition time [TR] = 1500 msec, echo time [TE] = 33 msec, voxel size = 2.0 mm isotropic, flip angle = 70°, multiband factor = 4, 52 slices automatically aligned to the AC-PC line). We acquired partial volumes fully covering the occipital and temporal lobes (see Appendix \ref{appendix-mri-acq}).

\subsection{3T fMRI Data Preprocessing}
In offline and real-time analyses, we analyzed the subject's data in native anatomical T1w-space.

\subsubsection{Offline Preprocessing} \label{offline-preprocessing}
Offline preprocessing was performed using fMRIPrep (Supplement \ref{fmriprep}). Single-trial responses (``\glslink{Betas}{betas}'') were estimated using GLMsingle \cite{princeImprovingAccuracySingletrial2022}, which uses repeated image presentations to improve single-trial response estimates. Repetitions are also used to determine reliable \glslink{Voxel}{voxels}, as described in Section \ref{reliability-mask}. Betas were z-scored voxelwise using the training images from the entire session. 

\subsubsection{Real-time Preprocessing}\label{rtpreproc}
Each \glslink{TR}{TR}, a 3-D brain volume is streamed from the MRI scanner (Appendix \ref{rt-infra}). At the beginning of the session, we compute an affine registration between the first functional volume and a \glslink{BOLD}{BOLD} reference volume from the offline training session using FSL (version 6.07.15; \citealp{jenkinsonFSL2012}) FLIRT (version 6.0) with 6 degrees of freedom. On all subsequent TRs, we use FSL MCFLIRT to compute a motion correction transformation for the current volume with reference to the first functional volume of the session. The two transformation matrices (for registration and motion correction) are then combined and applied to the data from the current TR to align the brain volume to the training scan. 

One challenge with real-time fMRI is that, when a participant views an image, the BOLD response is temporally extended, generally peaking $\sim$4-6 seconds after stimulus onset and not fully dissipating until $\sim$12–20 seconds. To address this, we delay inference until $\sim$7.9 seconds ($\sim$2 trials) after the onset of the target image to increase the likelihood of capturing the BOLD response peak (Appendix \ref{tr-labels}). This intrinsic temporal limitation of fMRI is discussed further in Limitations. The second challenge is that, due to this lag, the response to a given image may overlap with that of the next image when they are presented in rapid succession (as is the case here). To address this, we take a standard approach in fMRI data analysis, fitting a General Linear Model (\glslink{GLM}{GLM}) to account for the temporally-extended response to an image and extract a single-trial beta. We use the Least Squares-Separate implementation from Nilearn (version 0.11.1; \citealp{nilearn_contributors_2024_14546577,abrahamMachineLearningNeuroimaging2014,pedregosaScikitlearnMachineLearning2011}) which has been shown to work especially well for decoding analyses compared to other GLM approaches \cite{mumfordDeconvolvingBOLDActivation2012}. This results in a single beta vector for the to-be-decoded image, representing a ``current trial vs. all others'' comparison. Voxelwise betas are accumulated over the course of the entire session and z-scored with all data that are available at that time. 

\subsection{Pipeline Variations}
The temporal resolution of fMRI ($\sim$0.5-1 Hz) and the extended BOLD response (spanning $\sim$6-12 seconds) place an inherent lower bound on real-time decoding speed. As a result, we document results using different stimulus delays. Specifically, the ``delays'' refer to the amount of time we allow to elapse before attempting to fit the GLM (see Appendix \ref{tr-labels} for implementation). 

Here, we describe three main analysis speeds that fall under the umbrella of real-time compatible analysis. We define ``real-time compatible'' as an analysis that can be used to influence the experiment while the subject is still being scanned (e.g., adapting the stimuli shown to the subject based on some readout of brain activity or providing \glslink{Closed-loop Neurofeedback}{neurofeedback} to train the subject to modulate their neural activity). We call the three variations ``fast'', ``slow'', and ``end-of-run''. For ``fast'' real-time, we acquire functional data for $\sim$7.9 seconds ($\sim$2 trials) post stimulus-onset (Table \ref{table:latency}), which should be sufficient for the BOLD response for most voxels to peak. Preliminary explorations demonstrated that a lower amount of time is insufficient for decoding. For ``slow'' real-time, we acquire functional data for $\sim$29 seconds ($\sim$7 trials) post stimulus-onset, an intermediate point that optimizes the trade-off between delay and performance (Section \ref{results:3t-delay-vs-performance}). Finally, for ``end-of-run'' real-time, we acquire functional data until the end of the functional run -- roughly 5 minutes -- before fitting the GLM for each trial.  

\subsection{Model Architecture} \label{model-arch}
We used a condensed version of the MindEye2 architecture \cite{scottiMindEye2SharedSubjectModels2024}, omitting the low-level submodule, image-to-image refinement, and text caption refinement steps in addition to reducing the shared-subject latent dimensionality from $4096$ to $1024$. The authors of the original work reported that these changes minimally impacted evaluation metrics, but their removal lowers the parameter count and significantly speeds up training and inference. 

The inputs (voxelwise single-trial betas within a ``reliability mask''; see Section \ref{reliability-mask}) are converted to a shared-subject latent space using ridge regression \cite{hastieLinearMethodsRegression2009}. These latents are passed through a residual multilayer perceptron (MLP), resulting in an embedding with the same dimensionality as OpenCLIP ViT-bigG/14's image token embeddings (penultimate layer; $256 \times 1664$ dimensions) \cite{radfordLearningTransferableVisual2021}. This ``backbone'' embedding can then (depending on the desired application at inference-time) be passed to two separate branches: image reconstruction (generate an image) and image retrieval (select the seen image out of a pool of candidates). Despite this modularity, the whole model is trained end-to-end with a combined loss (see \citealp{scottiMindEye2SharedSubjectModels2024} for a comprehensive description).

For reconstruction, the backbone embedding is mapped to the CLIP model's image embedding space guided by a diffusion prior, and a frozen unCLIP model based on Stable Diffusion XL \cite{podellSDXLImprovingLatent2023} then generates the image reconstruction. For retrieval, the backbone embedding follows an analogous mapping to CLIP image space – but via an MLP ``projector'' rather than a diffusion prior. The resulting (predicted) fMRI-CLIP embedding can then be compared with the (ground-truth) CLIP image embeddings of candidate images; the top-$k$ retrievals are defined as the $k$-nearest neighbors (based on cosine similarity) to the predicted embedding.

\subsection{Model Training}
Prior to inference (offline or real-time), we follow the training procedure described by \citet{scottiMindEye2SharedSubjectModels2024}. We use a pretrained checkpoint trained on 7 of the 8 NSD subjects (\citealp{allenMassive7TFMRI2022}; subj01 is held-out from pretraining to replicate our main findings). Subsequently, we fine-tune the model on one session ($\sim$1 hour) of 3T data from a new participant (for replications, we use the held-out NSD subject). 

\subsubsection{Train and Test Split}
During pretraining, the model was trained on all images except the ``shared1000'', which were seen by all NSD participants (``special515'' refers to a subset of shared1000 for which all participants saw all three repetitions). For the 3T participant, the model was fine-tuned on all 543 images (481 unique) from the first scan except the 50 ``special515” images (repeated 3x each; 150 total). The model was then tested on three repeats of a different subset of 50 ``special515” images from the second (real-time) scan session. During pretraining and fine-tuning on 7T data, the inputs to the model were single-trial betas; fine-tuning on 3T data was done using betas averaged across all available repetitions of a given image in the first scan. 

\subsubsection{Selecting Reliable Voxels} \label{reliability-mask}
First, we transformed the ``nsdgeneral'' region of interest (ROI), a mask of voxels in occipital cortex that responded reliably in NSD, from MNI space into the subject's native anatomical T1w space. We then applied a subject-specific ``reliability mask'' on top of the nsdgeneral mask because this improved performance in some preliminary data. Reliability for a given voxel is defined as its average response correlation across repeated presentations of the same image. Specifically, we computed the Pearson's correlation of betas (from GLMsingle) across the first two instances of each repeated image in the training session and set a reliability threshold at $r>0.2$ to generate a binary mask.  

\subsection{Model Evaluation} \label{methods-evals}
The test set for all experiments (50 ``special515'' images, repeated 3x each) was kept identical to facilitate comparisons across variations of the analysis. Replications in NSD subj01 used the exact same test images. Unless otherwise stated, evaluations use single-trial betas from the first presentation of the three repeats only.

PixCorr (pixel correlation), SSIM (structural similarity index metric; \citealp{zhouwangImageQualityAssessment2004}), EfficientNet-B1 (“Eff”; \citealp{tanEfficientNetRethinkingModel2020}) and SwAV-ResNet50 (“SwAV”; \citealp{caronUnsupervisedLearningVisual2021}) compute the average correlation distance between the ground truth and reconstructed image. AlexNet (layers 2 and 5) \cite{krizhevskyImageNetClassificationDeep2012}, Inception-v3 (last pooling layer) \cite{szegedyRethinkingInceptionArchitecture2016}, and CLIP (last layer of ViT-L/14) \cite{radfordLearningTransferableVisual2021} use two-way comparisons (chance=50\%) based on extracted features from the specified layer. For each test image, we compute the Pearson's correlation of each ground truth image with each reconstruction in the test set; two-way accuracy refers to the number of times the correct ground-truth reconstruction pair is more correlated than a mismatched pair, averaged over all possible pairwise comparisons and then over all test images to produce a single score. 

We include two retrieval metrics: ``image'' (forward) retrieval and ``brain'' (backward) retrieval. Image retrieval uses the predicted CLIP embedding (based on fMRI) to choose the nearest neighbor ground-truth CLIP embedding. Brain retrieval is the opposite, using a ground-truth CLIP embedding to choose the nearest neighbor fMRI-CLIP embedding. The retrieval score in each case is the top-$1$ accuracy when repeating this procedure for each image in the test set (chance=1/50; 2\%).

\section{Results}
\subsection{Comparing Pipeline Variations}
\begin{figure}
  \centering
  \includegraphics[width=0.6\textwidth]{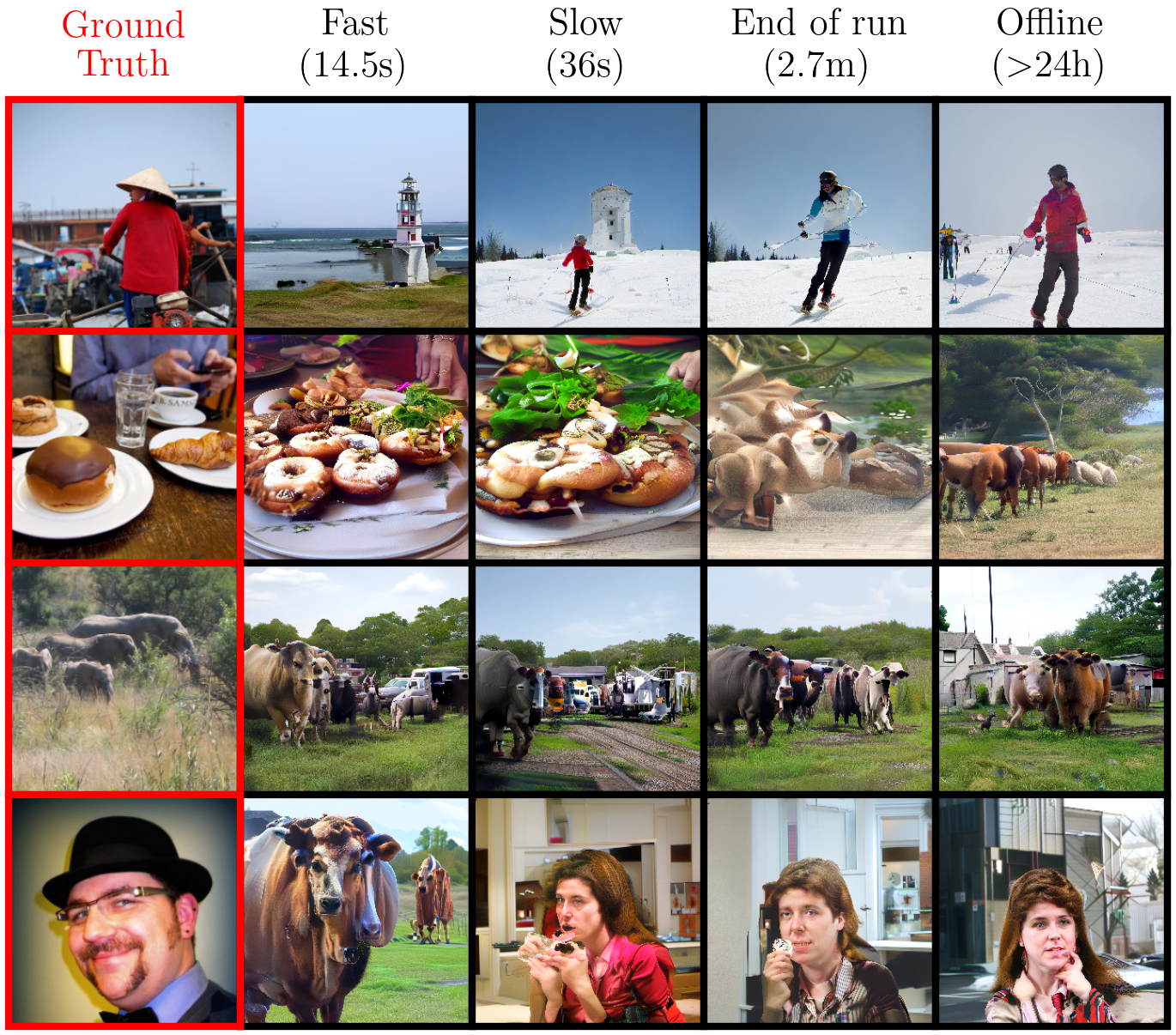}
  \caption{Hand-picked example reconstructions for different configurations in 3T.}
  \label{fig:example_recons}
\end{figure}

\begin{figure}
  \centering
  \includegraphics[width=0.8\textwidth]{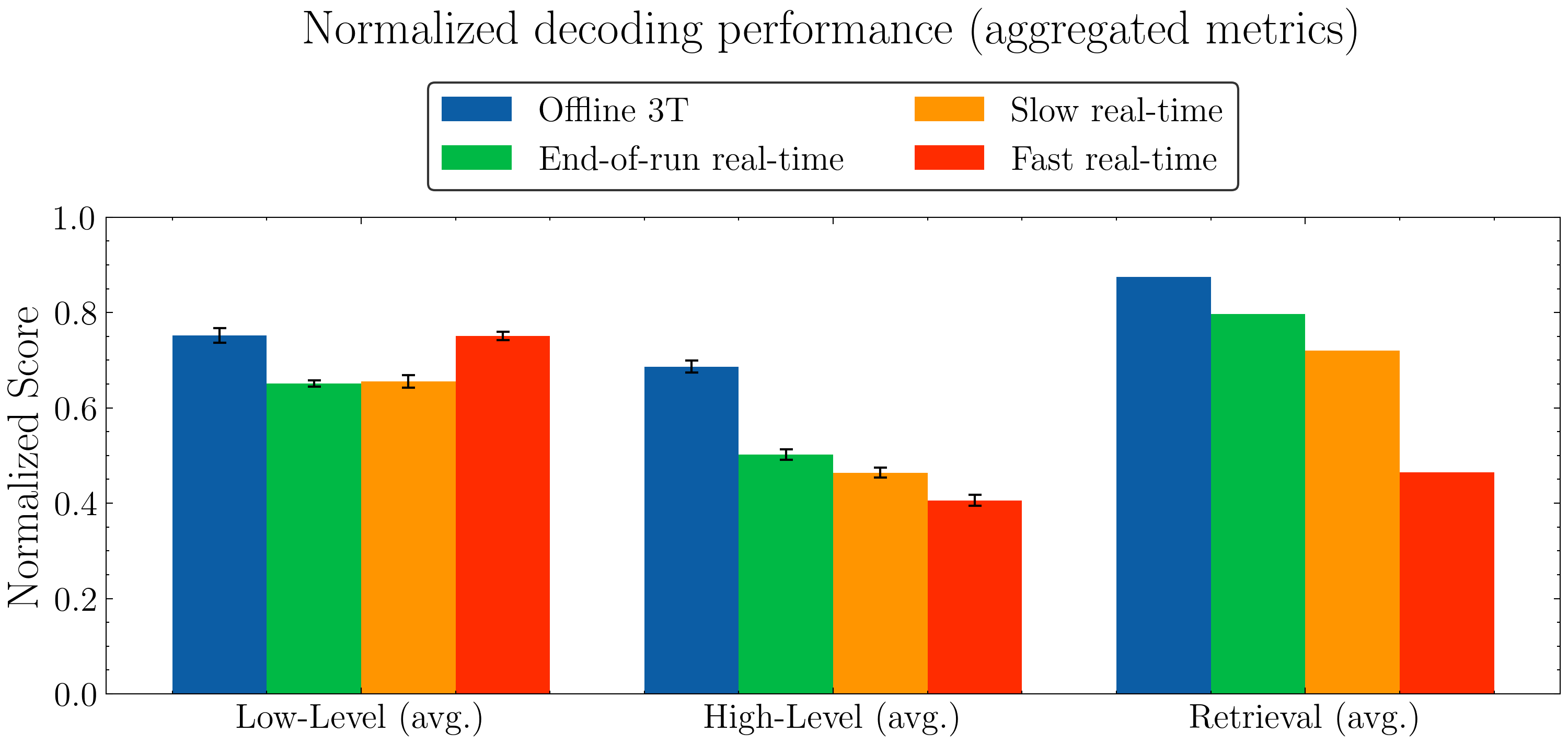}
  \caption{Aggregated evaluation metrics for 3T offline and real-time pipelines. Scores are min-max normalized per metric so that $0$ corresponds to using random COCO images as reconstructions and $1$ corresponds to offline NSD performance. Bars show the mean and standard error across 5 random seeds.}
  \label{fig:barplot_eval_aggregated}
\end{figure}

The performance of the offline 3T pipeline is close to offline 7T performance (Table \ref{tab:recon_eval}; Figure \ref{fig:barplot_eval_aggregated}). Previous work relied on 7T MRI; our results show that good offline fMRI-to-image performance is possible using a model that is pretrained on 7T data and then fine-tuned and tested on 3T data (Figure \ref{fig:example_recons}; see Figure \ref{fig:large_grid_3T} for more examples). Note that the 7T and 3T tests were conducted on different participants, so we caution against attributing the difference in performance exclusively to scanner field strength.

\begin{table*}[htbp]
    \centering
    \setlength{\tabcolsep}{.9pt}
    \footnotesize
    \begin{tabular}{lccccccccccc}
        \toprule
        Method & Latency & \multicolumn{4}{c}{Low-Level} & \multicolumn{4}{c}{High-Level} & \multicolumn{2}{c}{Top-$1$ Retrieval}\\
        \cmidrule(lr){3-6} \cmidrule(lr){7-10} \cmidrule(l){11-12}
         & & PixCorr $\uparrow$ & SSIM $\uparrow$ & Alex(2) $\uparrow$ & Alex(5) $\uparrow$ & Incep $\uparrow$ & CLIP $\uparrow$  & Eff $\downarrow$ & SwAV $\downarrow$ & Image $\uparrow$ & Brain $\uparrow$\\
        \midrule
        Offline 7T (avg. 3 reps.) & $1$d & $0.231$ & $0.349$ & $88.4\%$ & $95.6\%$ & $85.9\%$ & $78.8\%$ & $0.803$ & $0.423$ & $100\%$ & $96\%$\\
        Offline 3T (avg. 3 reps.) & $1$d & $0.160$ & $0.343$ & $86.6\%$ & $91.1\%$ & $75.9\%$ & $75.4\%$ & $0.854$ & $0.493$ & $90\%$ & $88\%$\\
        \midrule
        Offline 7T & $1$d & $0.228$ & $0.330$ & $84.5\%$ & $93.1\%$ & $85.5\%$ & $78.5\%$ & $0.832$ & $0.448$ & $78\%$ & $82\%$ \\
        Offline 3T & $1$d & $0.101$ & $0.328$ & $79.7\%$ & $83.8\%$ & $73.2\%$ & $72.1\%$ & $0.884$ & $0.516$ & $76\%$ & $64\%$ \\
        End-of-run real-time & $2.7$m& $0.052$ & $0.330$ & $75.4\%$ & $79.8\%$ & $68.9\%$ & $65.0\%$ & $0.918$ & $0.544$ & $66\%$ & $62\%$ \\
        Slow real-time & $36$s& $0.062$ & $0.334$ & $73.8\%$ & $77.3\%$ & $68.6\%$ & $63.7\%$ & $0.928$ & $0.550$ & $58\%$ & $58\%$ \\
        Fast real-time & $14.5$s& $0.088$ & $0.350$ & $74.0\%$ & $75.6\%$ & $64.6\%$ & $62.9\%$ & $0.926$ & $0.572$ & $36\%$ & $40\%$ \\
        \bottomrule
    \end{tabular}
    \caption{Latency and reconstruction/retrieval metrics for 3T and 7T offline and real-time pipelines. Reconstruction metrics are averaged over 5 random seeds; retrieval is deterministic (see Section \ref{methods-evals} for details on computing metrics). $\uparrow$ ($\downarrow$) means higher (lower) scores are better.}
    \label{tab:recon_eval}
\end{table*}

As expected, fully offline analyses perform better than their real-time compatible counterparts across most evaluation metrics (Table \ref{tab:recon_eval}, Figure \ref{fig:barplot_eval_aggregated}). The performance gap between the offline and end-of-run pipelines suggests that the inclusion of extensive preprocessing steps such as fMRIPrep and GLMsingle in the offline pipeline is beneficial, but not essential, for single-trial decoding. Importantly, even the fastest real-time analysis performs above chance-level (this is true even without pretraining on NSD; Appendix \ref{appendix-nopretrain}).  We qualitatively replicate these results using the held-out NSD subj01 (Appendix \ref{appendix-nsd-rt}). As noted by \citet{scottiMindEye2SharedSubjectModels2024}, SSIM substantially diverges from the other metrics (Figure \ref{fig:barplot_eval_full}), suggesting that this metric may not be an accurate reflection of reconstruction quality.

\begin{figure}
\centering
\begin{subfigure}{.5\textwidth}
  \centering
  \includegraphics[width=\linewidth]{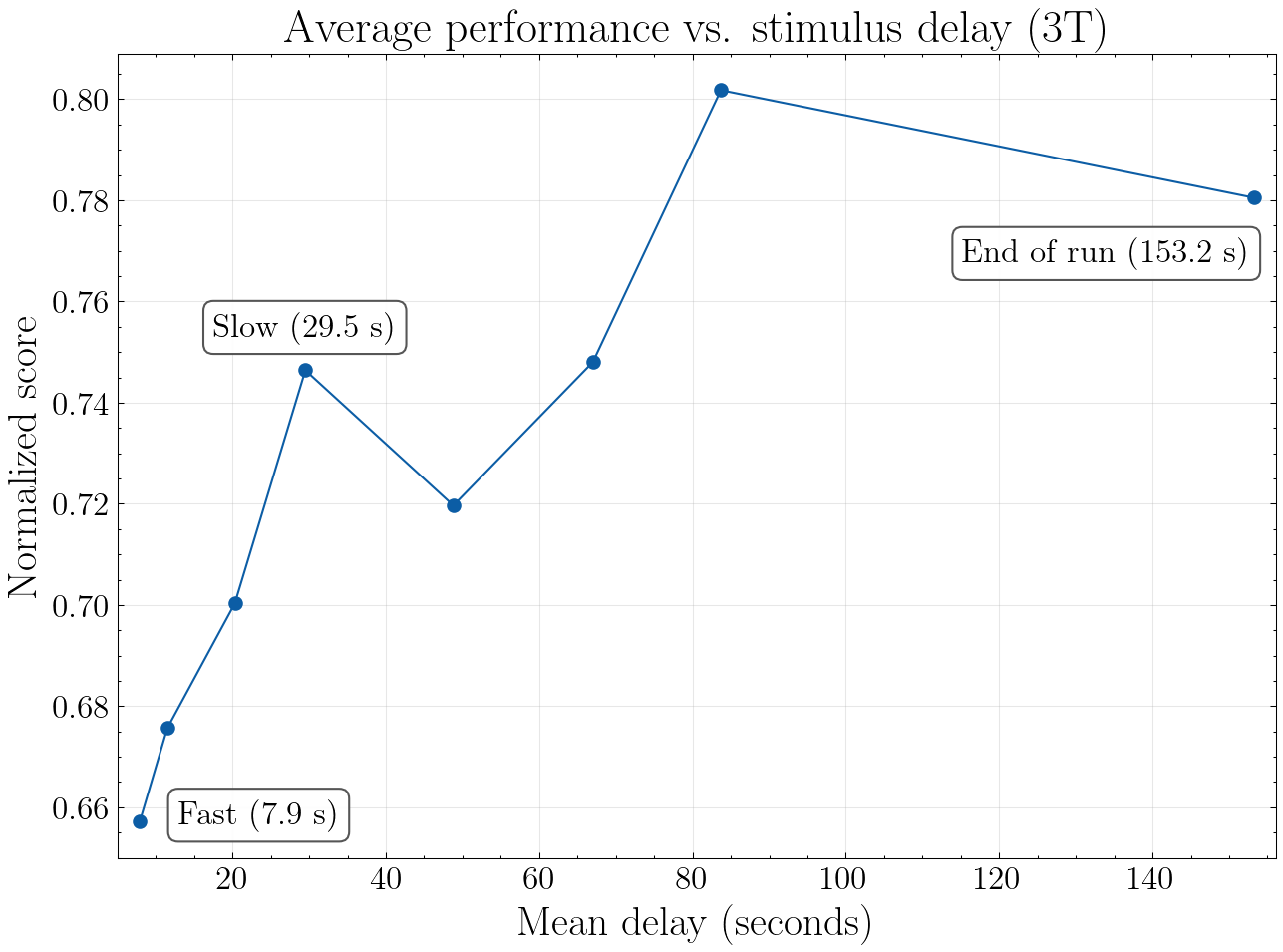}
  \caption{}
  \label{fig:3t-delay-vs-performance}
\end{subfigure}%
\begin{subfigure}{.5\textwidth}
  \centering
  \includegraphics[width=\linewidth]{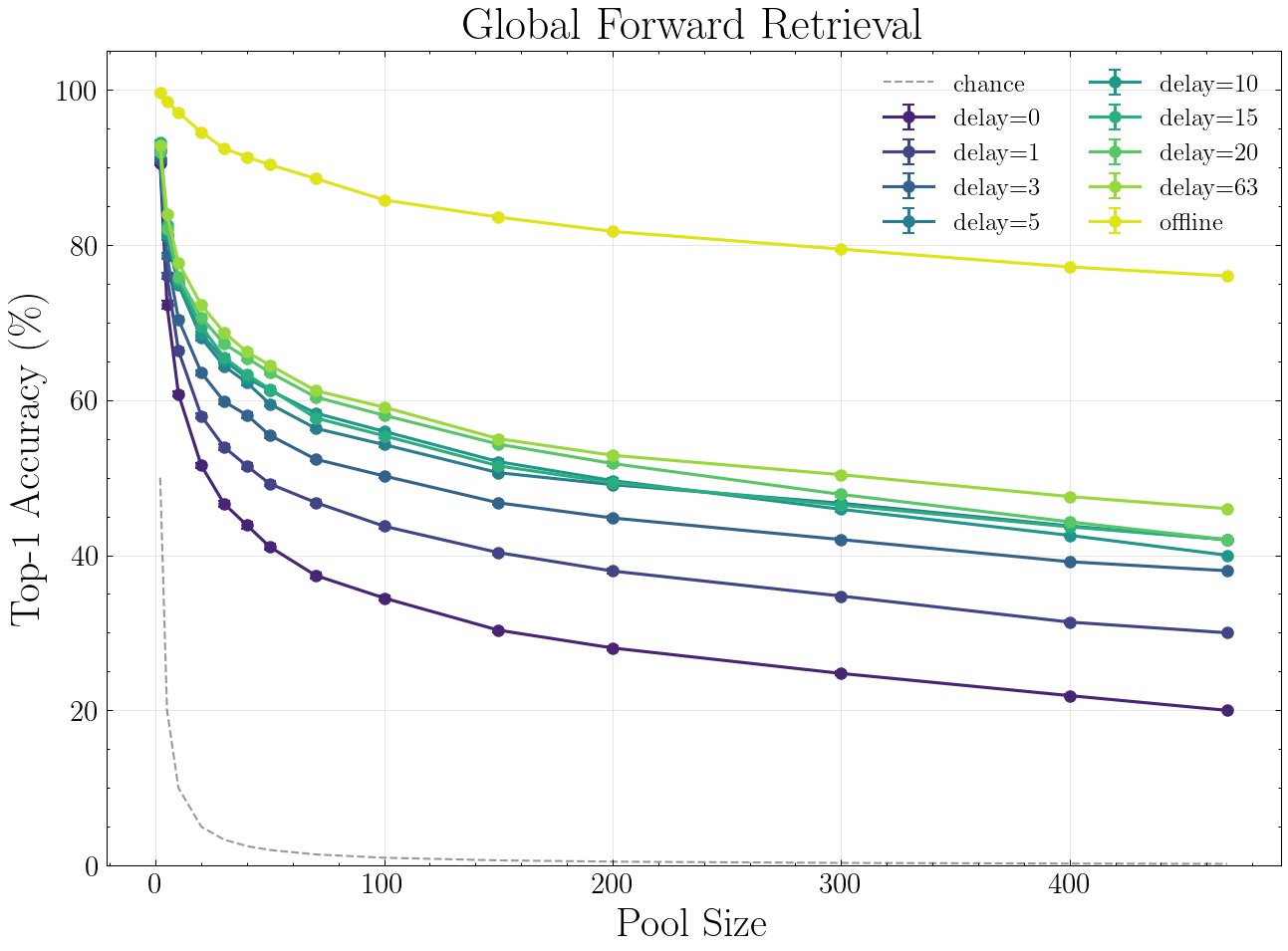}
  \caption{}
  \label{fig:global-retrieval}
\end{subfigure}
\caption{Relationships between stimulus delay, decoding performance, and retrieval pool size. (a) Performance (averaged across reconstruction/retrieval metrics listed in Table 1) vs. delays. Scores are min-max normalized per metric across delays, where 0 corresponds to using random COCO images as reconstructions and 1 corresponds to offline NSD. (b) Image retrieval across delays for varying pool sizes. Delay 0 = fast, delay 5 = slow, delay 63 = end-of-run.}
\label{fig:test}
\end{figure}

\subsection{Stimulus Delay and Retrieval Pool Size vs. Performance} \label{results:3t-delay-vs-performance}
There is a positive relationship between stimulus delay (i.e., how long we wait before starting to analyze the response to a stimulus) and decoding performance, implying that collecting additional data before fitting the GLM improves the quality of the response estimates (Figure \ref{fig:3t-delay-vs-performance}). This relationship has an elbow at roughly 30 seconds before diminishing returns, which suggests that this may be an optimal trade-off point between speed and accuracy. We replicated this pattern of findings by implementing our real-time compatible pipeline using NSD subj01 (Appendix \ref{appendix-nsd-delay-vs-performance}). 

Figure \ref{fig:global-retrieval} shows top-$1$ retrieval accuracy as a function of candidate pool size and stimulus delay. Higher delay values show more robustness to increasing pool size (i.e., there is less of a decline in performance as pool size increases). Notably, all real-time variants perform similarly at the smallest pool size (i.e., size 2; see Section \ref{discussion} for discussion) and all points are above chance.

\begin{table}
    \centering
    \setlength{\abovecaptionskip}{5pt}
    \setlength{\tabcolsep}{1pt}
    \footnotesize
    \begin{tabular}{|l|ccc|}
    \hline
    &Fast (s)&Slow (s)&End-of-run (s)\\
   \hline
    \emph{Stimulus Delay}& $7.85 {\scriptstyle \pm 0.59}$& $29.45 {\scriptstyle \pm 2.63}$& $153.18 {\scriptstyle \pm 79.63}$\\
    Motion Correction            & $0.39 {\scriptstyle \pm 0.01}$& $0.39 {\scriptstyle \pm 0.01}$& $0.39 {\scriptstyle \pm 0.01}$\\
    Registration                 & $0.18 {\scriptstyle \pm 0.00}$& $0.18 {\scriptstyle \pm 0.00}$& $0.18 {\scriptstyle \pm 0.00}$\\
    \emph{GLM Fit}& $1.09 {\scriptstyle \pm 0.12}$& $1.11 {\scriptstyle \pm 0.11}$& $1.21{\scriptstyle \pm 0.08}$\\
    Inference& $0.19 {\scriptstyle \pm 0.01}$& $0.19 {\scriptstyle \pm 0.01}$& $0.19 {\scriptstyle \pm 0.01}$\\
    Reconstruction& $4.56 {\scriptstyle \pm 0.05}$& $4.56 {\scriptstyle \pm 0.05}$& $4.56 {\scriptstyle \pm 0.05}$\\
    Retrieval& $0.50 {\scriptstyle \pm 0.00}$& $0.50 {\scriptstyle \pm 0.00}$& $0.50 {\scriptstyle \pm 0.00}$\\  
    \hline
    Total Latency                & $14.76 {\scriptstyle \pm 0.62}$& $36.38 {\scriptstyle \pm 2.65}$& $160.21 {\scriptstyle \pm 79.65}$\\
    \hline
    \end{tabular}
\caption{Time for real-time fMRI preprocessing and MindEye2 inference (mean ± standard deviation over all session trials). Steps where processing time differs across variations are \emph{italicized}.} 
\label{table:latency}
\end{table}

\subsection{Real-time Analysis Latency}
The amount of time required for motion correction, registration, inference, reconstruction, and retrieval is the same for all real-time processing variants (Table \ref{table:latency}). This is because the same operations are being executed regardless of the stimulus delay, and the input data are always the same size. Fitting the GLM is slightly slower for longer delays because there is more functional data present per trial as the delay grows. 

Including the pre-defined stimulus delay (the delay between stimulus onset and the start of analysis in order to wait for the BOLD response), ``fast'' retrieval takes $\sim$10s from stimulus onset and ``fast'' reconstruction takes $\sim$14s. It takes $\sim$15s to perform both components (Table \ref{table:latency}; a majority of this time is due to the stimulus delay and the diffusion prior for reconstruction). 

\section{Discussion} \label{discussion}
\subsection{The Transition from Offline to Real-time}
\label{discussion-offline-vs-realtime}
Table \ref{tab:recon_eval} demonstrates the large impact of averaging repetitions on decoding accuracy. The majority of prior works average responses across stimulus repetitions to increase signal-to-noise; for example, MindEye2 uses three repeated presentations. Related works using magnetoencephalography (MEG; \citealp{benchetritBrainDecodingRealtime2024}) and electroencephalography (EEG; \citealp{kneelandENIGMAUnifiedLightweight2025}) (which have better temporal but worse spatial resolution than fMRI) rely on 12 and even up to 80 repetitions of the same image to extract a reliable signal. Because our focus is on real-time applications, we use only the first presentation of each test image. While there is a meaningful decrease in performance going from three presentations to one presentation, it is promising that decoding performance remains strong when there is no averaging across stimulus repetitions.

Furthermore, the positive relationship between processing delay and decoding performance is encouraging, suggesting that -- depending on the use-case -- a researcher may wish to select a particular point along the trade-off continuum between real-time decoding speed and accuracy. There are two potential reasons that performance increases at latencies longer than the duration of the BOLD response to the trial of interest. First, the \glslink{GLM}{GLM} approach to response estimation models the hemodynamic response of a \glslink{Voxel}{voxel} to the trial of interest while accounting for its response to other seen images; including more images may be beneficial as the GLM is able to gain more information related to a voxel's \emph{differential selectivity} for an image. Second, because the GLM at longer stimulus delays is fit using more data (in general), the model may do better at estimating confounds like temporal drift in the BOLD signal. Future work can more precisely identify how these factors contribute to improved performance. 


\subsection{Towards Non-Invasive Brain-Computer Interfaces}
The core aim of this family of generative approaches to fMRI-to-image reconstruction \cite{ozcelikReconstructionPerceivedImages2022,ozcelikNaturalSceneReconstruction2023a,scottiReconstructingMindsEye2023a,scottiMindEye2SharedSubjectModels2024,beliyBrainITImageReconstruction2025} is to learn a mapping from fMRI responses to the latent embedding space of a pretrained foundation model. In this case, we use CLIP \cite{radfordLearningTransferableVisual2021}, which is jointly trained on large-scale image and language data, resulting in a rich, visual-semantic latent space. Mapping brain activity into this feature space enables a range of real-time applications including fMRI-to-image reconstruction. Importantly, there are also promising applications of this approach that operate directly on the predicted latent embeddings, without the need to reconstruct. For example, to get participants to represent two images in a more similar fashion (i.e., to integrate them), participants could be rewarded when fMRI scans corresponding to those images are mapped to coordinates that are closer together in the latent space (for another approach to fostering integration with \glslink{Closed-loop Neurofeedback}{neurofeedback}, see \cite{pengInducingRepresentationalChange2024}). Conversely, to get participants to represent two images more distinctly (i.e., to differentiate them), participants could be rewarded when fMRI scans corresponding to those images are mapped to coordinates that are further apart in the latent space.

The usefulness of the pipeline described here for brain-computer interfaces depends critically on both the \emph{accuracy} of the mapping from fMRI data into latent space and also the \emph{speed} with which results can be returned. With regard to accuracy: While there is still clearly room for improvement relative to offline methods, it is encouraging that single-trial decoding accuracy is still well above-chance even at the lowest latencies (Figure \ref{fig:global-retrieval}) -- in particular, the strong retrieval accuracy (90.6\%) at pool size=2 in the ``fast'' real-time setting suggests that our pipeline, in its present form, may be able to support neurofeedback based on relative distances in latent space (e.g., encouraging participants to shift their representations of images to be closer together or farther apart, as discussed above). With regard to speed: The required speed-of-decoding will of course depend on the particular brain-computer interface application. In neurofeedback setups, longer lags create a temporal credit assignment problem where it becomes difficult for the participant to match the feedback to the cognitive state that triggered the feedback, impeding learning \cite{lubianikerNeurofeedbackLensReinforcement2022a,sitaramClosedloopBrainTraining2017}. 
Fortunately, extant studies provide strong evidence that the decoding latencies that we explore here (ranging from 10-30s) are suitable for learning -- indeed, studies have reported promising learning effects with even slower (20-40s) latencies \cite{ganesanNeurofeedbackTrainingFacilitates2025,lubianikerUpregulationRewardMesolimbic2026}. \citet{oblakSelfregulationStrategyFeedback2017} used simulations to show that, in certain conditions, intermittent feedback may actually drive learning better than continuous feedback, especially under realistic assumptions regarding the blurred and delayed hemodynamic response of fMRI. Importantly, these studies (and most prior neurofeedback work, with some notable exceptions such as \citealp{buschHumanLearningNoninvasive2026}) delivered feedback based on simple neural signals. Our framework can deliver much more fine-grained feedback about the participant's cognitive state within the same time budget. 

\section{Future Directions}
\citet{banvilleScalingLawsDecoding2025} demonstrated that image decoding performance is primarily driven by the amount of data per subject, as opposed to the total number of subjects. In line with these findings, we demonstrate improved real-time decoding performance with one additional fine-tuning session in 3T and replicate this using more sessions from NSD (Appendix \ref{appendix-metrics-vs-training-data}). Future work may explore this by performing multiple training sessions prior to real-time inference in order to maximize decoding performance. Additionally, active learning approaches (also known as adaptive stimulus optimization), which select the most informative training images from a larger pool of candidate images, can be used to maximize the performance benefits that are obtained from a given (limited) number of training sessions \cite{dimattinaAdaptiveStimulusOptimization2013,cowleyAdaptiveStimulusSelection2017,jhaActiveLearningDiscrete2024}. 

Other analysis pipelines and architectures may replace the approaches used here; for example, \citet{careilDynadiffSinglestageDecoding2025} have demonstrated a method for time-resolved decoding from fMRI and \citet{beliyBrainITImageReconstruction2025} have shown better transfer learning in limited data settings. Improved pipelines can supersede the preprocessing steps used here or the MindEye2 architecture and can easily be integrated with the RT-Cloud framework. 

In addition, the field may increasingly benefit from the incorporation of fMRI foundation models \cite{caroBrainLMFoundationModel2023,kimSwiFTSwin4D2023,dongBrainJEPABrainDynamics2024,dongBrainHarmonyMultimodal2025,gijsenBrainSemantoksLearningSemantic2026,laneScalingVisionTransformers2026}: self-supervised pretraining approaches that leverage large quantities of unlabeled data; these foundation models can then be fine-tuned to a specific task using a smaller amount of labeled data, improving performance \cite{azabouUnifiedScalableFramework2023b,zhangNeuralEncodingDecoding2025,ryooGeneralizableRealtimeNeural2025}. Such approaches may further extend the benefits of cross-subject pretraining demonstrated in \citet{scottiMindEye2SharedSubjectModels2024} that we use here, potentially reducing the amount of data required from a new participant and increasing the ceiling on decoding performance.

\section{Conclusion} \label{conclusion}
Here, we describe a real-time compatible adaptation of MindEye2, a state-of-the-art machine learning pipeline that can decode seen images from fMRI by mapping fMRI responses to the latent space of a language-image foundation model (CLIP). After developing these methods, we performed a proof-of-concept real-time fMRI scan that implemented real-time MindEye using the RT-Cloud framework \cite{wallaceRTCloudCloudbasedSoftware2022a}. In this manuscript, we report results from simulated real-time analyses with varying preprocessing pipelines to document the factors driving changes in performance from offline to real-time analysis. Beyond the image decoding use-case demonstrated here, this work demonstrates how computationally intensive AI algorithms can be implemented within RT-Cloud to support new kinds of real-time fMRI analysis. To encourage future work that employs such approaches to implement novel closed-loop experiments and non-invasive brain-computer interface technologies, we provide open-source data and code, and we are continuing to develop these methods in the open; we extend a global invitation for interested parties to collaborate with us on developing new extensions or applications of this work.

\section*{Software and Data}
Code used for real-time analysis is located on GitHub at \url{https://github.com/brainiak/rtcloud-projects/tree/main/mindeye}. We include documentation and tutorials to replicate the real-time pipeline (both in simulation and on an fMRI scanner using RT-Cloud) as well as scripts for offline fMRI preprocessing, analysis, and model training. Offline-preprocessed fMRI data collected at Princeton can be found on HuggingFace at \url{https://huggingface.co/datasets/rishab-iyer1/fmriprep_mindeye}. GLMsingle betas derived from the offline-preprocessed Princeton data can be found at \url{https://huggingface.co/datasets/rishab-iyer1/glmsingle/tree/main}. Data for real-time fMRI can be found at \url{https://huggingface.co/datasets/rishab-iyer1/3t}.

\begin{ack}
This work benefited greatly from the MedARC (Medical AI Research Center) Discord community\footnote{\url{https://discord.com/invite/tVR4TWnRM9}}, an open-source forum for conducting research on topics related to neuroscience, medicine, and AI. We thank Sophont for organizing the MedARC community and supporting open-science research by providing access to GPU resources for volunteers. We thank the members of the Princeton Computational Memory Lab for helpful discussions and feedback, including Augustin Hennings, Nitzan Lubianiker, Wanjia Guo, Samuel Nastase, Cody Dong, Ariadne Letrou, and Alex Nguyen. We thank the Scully Center team -- Mark Pinsk, Nicholas DePinto, and Leigh Nystrom -- at the Princeton Neuroscience Institute for technical assistance with fMRI scanning and real-time fMRI. We thank the Princeton Neuroscience Institute's Research Computing and Scientific Computing staff -- Garrett McGrath and Benjamin Singer -- for assistance setting up a GPU workstation for real-time fMRI. We thank the contributors to open-source Python libraries for scientific computing, machine learning, and neuroimaging, including NumPy \cite{numpy}, Scipy \cite{scipy}, Matplotlib \cite{Matplotlib}, pandas \cite{reback2020pandas}, HuggingFace Transformers \cite{transformers}, PyTorch \cite{pytorch}, Nilearn \cite{nilearn_contributors_2024_14546577}, and Nibabel \cite{brett_2025_17833216}, without which this work would not be possible. 

Funding was provided by the Princeton University Dean for Research Innovation Fund for New Industrial Collaborations and Sophont.  Early stages of this work (before the involvement of the MedARC community) were supported by NIH grant RF1MH125318. Funding for the acquisition of the data was provided in part by the Regina and John Scully ’66 Center for the Neuroscience of Mind and Behavior at Princeton University. 
\end{ack}

\clearpage

\defaultbibliographystyle{unsrtnat}
\defaultbibliography{rt-mindeye}
\putbib[rt-mindeye,fmriprep]
\clearpage

\appendix
\section{Appendix}
\subsection{Limitations}
\label{limitations}
Functional MRI has relatively poor temporal resolution compared with other non-invasive methods (EEG and MEG) and invasive methods, which can resolve neural activity on the order of milliseconds. This inherent temporal limitation places a lower bound on response latency for closed-loop feedback and brain-computer interface applications, though fMRI remains the most spatially resolved non-invasive brain imaging tool, motivating its use here. FMRI is also not portable, requiring participants to visit an MRI facility (generally located at a university or a hospital). Additionally, repeated or long-term interventions are difficult, requiring multiple visits to the scanner; on the other hand, invasive brain-computer interfaces are implanted once (often with neurosurgery) and can subsequently record data on a continuous basis for the lifespan of the device \cite{pelsStabilityChronicImplanted2019,davis5YearFollowupFully2025, mitchellAssessmentSafetyFully2023}. 

One limitation of our 7T pretraining procedure is that the images used for pretraining were interleaved with some of the images that were later (in a separate session) used for testing. Due to the temporal lag of the BOLD response, this could have led to a minor form of data leakage, whereby the neural response to the test images affects the beta maps for images presented after them during pretraining; this issue could be addressed in future work by defining a train/test split during pretraining based on continuous blocks of time as in \citet{careilDynadiffSinglestageDecoding2025}. Importantly, we expect that this issue would not significantly inflate our results, given that the model was only exposed to the fMRI data from the test images, but not the labels (CLIP embeddings) for these images. Our use of a fully held-out session for evaluation further minimizes this risk. 

We view the results presented here as a proof-of-concept demonstration that real-time decoding of individual stimuli from fMRI is feasible with limited training data. Due to our limited sample size, we cannot make claims about generalization to a broader population. Furthermore, we expect that decoding performance for the real-time session will be largely dependent on the similarity between the data used for model training and the data used for real-time testing. For example, some important factors to consider include the overlap in stimulus distribution between training and testing (see \cite{shirakawaSpuriousReconstructionBrain2025} for detailed analysis and critiques), the duration for which the stimulus is presented, and the task that the participant is performing. 

Finally, while the reconstructions reported here are well above chance on quantitative metrics, they clearly lack a high degree of fidelity with respect to the ground-truth image. While a long-term goal of the field is to be able to use perceptual decoders as a window into someone's subjective interpretation of an image, we are far away from that point and caution against over-interpretations of the generated reconstructions as providing any meaningful insight into internal cognitive processes. 

\subsection{Impact Statement}
In this paper, we have shown that it is possible to reliably decode visual perception from fMRI as early as a participant's second visit, within $\sim$10 seconds after they view an image. Such fine-grained decoding promises a window into people's ongoing cognitive processes as they unfold, and opens the door to several novel lines of research including closed-loop experimental paradigms and brain-computer interface (BCI) technologies. In addition to applications in scientific discovery, we hope that these methods may advance clinical treatments.

As the field continues to develop methods to decode participants' internal states in pursuit of these goals, it is imperative that researchers and companies follow ethical research practices, especially protecting participants' behavioral and neural data and preserving their privacy and autonomy. This includes, at minimum, informed consent from participants and considerations of the kinds of mental content that may be decodable. These values are captured by the principles of Respect for Persons, Beneficence, and Justice \cite{nationalcommissionfortheprotectionofhumansubjectsofbiomedicalandbehavioralresearchBelmontReportEthical1979}, to which researchers have a professional, ethical, and moral obligation.

\subsection{Real-time fMRI Infrastructure and Compute}
\label{rt-infra}
The MRI reconstruction computer reconstructs DICOM images given the raw k-space data from the scanner. These DICOMs are sent to a console host in the MRI control room (running Siemens syngo.MR software) via gigabit Ethernet. For real-time scans, we enable the DICOMs to be forwarded to a Linux machine by mounting a directory from that machine onto the console host using SMB. The Linux machine runs RT-Cloud which performs DICOM to NIFTI conversion as well as all of the real-time fMRI data preprocessing and analysis. The console host and the Linux machine are connected on a dedicated VLAN with a gigabit switch. 

The real-time analysis and most development was performed on a Linux system with a NVIDIA RTX 6000 Ada Generation GPU (49GB VRAM). Various training runs and experiments (some of which are not included in the paper) used one A100 80GB GPU node, one H100 80GB GPU node, and one 4xH100 80GB GPU node.

\subsection{Fast Real-time (No Pretraining)}
\label{appendix-nopretrain}
Table \ref{tab:recon_eval_nopretrain} shows reconstruction and retrieval evaluation metrics for the 3T real-time session without pretraining on NSD. We include a chance-level baseline corresponding to using randomly selected COCO images as the ``reconstructions''. Even without pretraining on NSD (i.e., only training the model with one session of 3T fMRI data), performance is still above this baseline, demonstrating that a large pretraining dataset (NSD) is not necessary for above-chance real-time decoding. 
\begin{table*}[htbp]
    \centering
    \captionsetup{font=small}
    \setlength{\tabcolsep}{1pt}
    \small
    \resizebox{0.95\columnwidth}{!}{

    \begin{tabular}{lccccccccccc}
        \toprule
        Method & Latency & \multicolumn{4}{c}{Low-Level} & \multicolumn{4}{c}{High-Level} & \multicolumn{2}{c}{Retrieval}\\
        \cmidrule(lr){3-6} \cmidrule(lr){7-10} \cmidrule(l){11-12}
         & & PixCorr $\uparrow$ & SSIM $\uparrow$ & Alex(2) $\uparrow$ & Alex(5) $\uparrow$ & Incep $\uparrow$ & CLIP $\uparrow$  & Eff $\downarrow$ & SwAV $\downarrow$ & Image $\uparrow$ & Brain $\uparrow$\\
        \midrule
        Fast real-time & $14.5$s& 0.065& 0.401& $62.61\%$& $63.88\%$& $60.20\%$& $56.82\%$& $0.935$& $0.580$& $22\%$& $22\%$\\
        Random Baseline& -& 0.014& 0.277& $50.25\%$& $50.99\%$& $50.37\%$& $50.38\%$& $0.982$& $0.655$& $1.6\%$& $1.4\%$\\
        \bottomrule
    \end{tabular}}
    \caption{Latency and reconstruction/retrieval evaluation metrics on the 3T fast real-time pipeline without pretraining on NSD data. $\uparrow$ ($\downarrow$) means higher (lower) scores are better. The baseline uses randomly chosen COCO images (excluding the ``shared1000'' images) as ``reconstructions'' for each test image.}
    \label{tab:recon_eval_nopretrain}
\end{table*}
\subsection{Replicating Real-time Pipelines using NSD}\label{appendix-nsd-rt}
\begin{figure}[ht]
  \centering
  \includegraphics[width=0.7\textwidth]{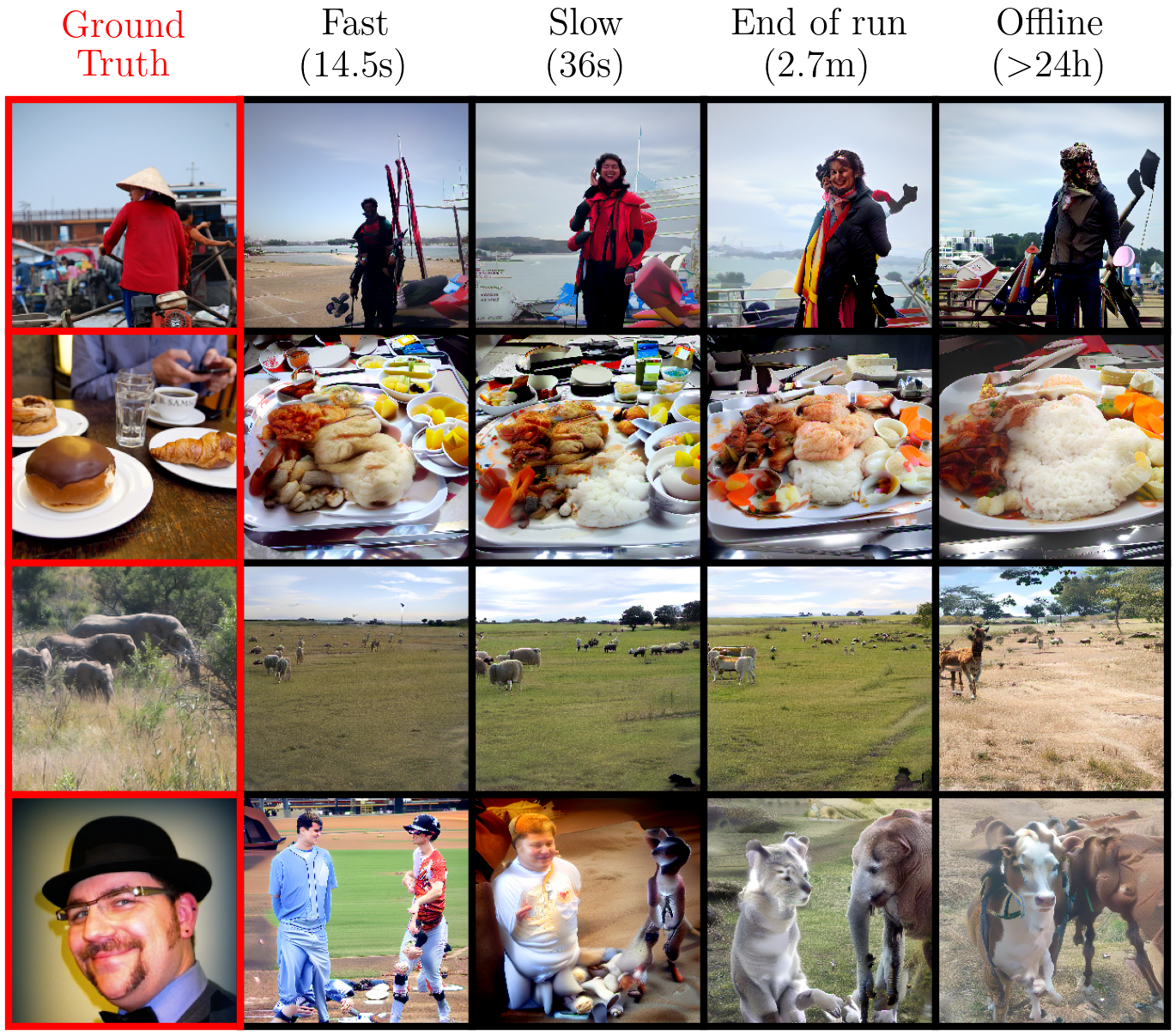}
  \caption{Hand-picked example reconstructions for different configurations on NSD subject 1.}
  \label{fig:NSD_example_recons}
\end{figure}

Here, we used the same pretrained checkpoint as all other 3T and 7T experiments and then fine-tuned the model using the first session of fMRI data from NSD subj01. Differences in procedures include the number of training images per session (693 in 3T and 750 in NSD) and masking (the nsdgeneral mask was applied without an additional subject-specific reliability mask). The test trials consisted of the first presentation of the exact same subset of 50 special515 images used in the 3T data; however, these trials spanned multiple sessions rather than a single session. The results (Table \ref{tab:recon_eval_NSD}, Figure \ref{fig:NSD_example_recons}, Figure \ref{fig:NSD-delay-vs-performance}) show a similar qualitative pattern of results to Table \ref{tab:recon_eval} where decoding performance improves with latency.

\begin{table*}[htbp]
    \centering
    \captionsetup{font=small}
    \setlength{\tabcolsep}{1pt}
    \small
    \resizebox{0.95\columnwidth}{!}{

    \begin{tabular}{lccccccccccc}
        \toprule
        Method & Latency & \multicolumn{4}{c}{Low-Level} & \multicolumn{4}{c}{High-Level} & \multicolumn{2}{c}{Retrieval}\\
        \cmidrule(lr){3-6} \cmidrule(lr){7-10} \cmidrule(l){11-12}
         & & PixCorr $\uparrow$ & SSIM $\uparrow$ & Alex(2) $\uparrow$ & Alex(5) $\uparrow$ & Incep $\uparrow$ & CLIP $\uparrow$  & Eff $\downarrow$ & SwAV $\downarrow$ & Image $\uparrow$ & Brain $\uparrow$\\
        \midrule
        Offline & $1$d & $0.228$ & $0.330$ & $84.5\%$ & $93.1\%$ & $85.5\%$ & $78.5\%$ & $0.832$ & $0.448$ & $78\%$ & $82\%$ \\
        End-of-run real-time & $2.7$m& $0.153$ & $0.335$ & $78.4\%$ & $85.3\%$ & $73.9\%$ & $67.4\%$ & $0.882$ & $0.492$ & $66\%$ & $62\%$ \\
        Slow real-time & $36$s& $0.174$ & $0.333$ & $77.2\%$ & $82.6\%$ & $69.9\%$ & $65.2\%$ & $0.897$ & $0.505$ & $58\%$ & $58\%$ \\
        Fast real-time & $14.5$s& $0.133$ & $0.313$ & $71.3\%$ & $76.5\%$ & $67.3\%$ & $63.7\%$ & $0.918$ & $0.533$ & $36\%$ & $40\%$ \\
        \bottomrule
    \end{tabular}}
    \caption{Summary of latency and reconstruction/retrieval evaluation metrics for offline and real-time decoding pipelines using NSD subject 1. Reconstruction metrics are averaged over 5 random seeds; retrieval is deterministic. $\uparrow$ ($\downarrow$) means higher (lower) scores are better.}
    \label{tab:recon_eval_NSD}
\end{table*}

\FloatBarrier

\subsection{Replicating Stimulus Delay vs. Performance using NSD} \label{appendix-nsd-delay-vs-performance}
\begin{figure}
  \centering
  \includegraphics[width=0.7\textwidth]{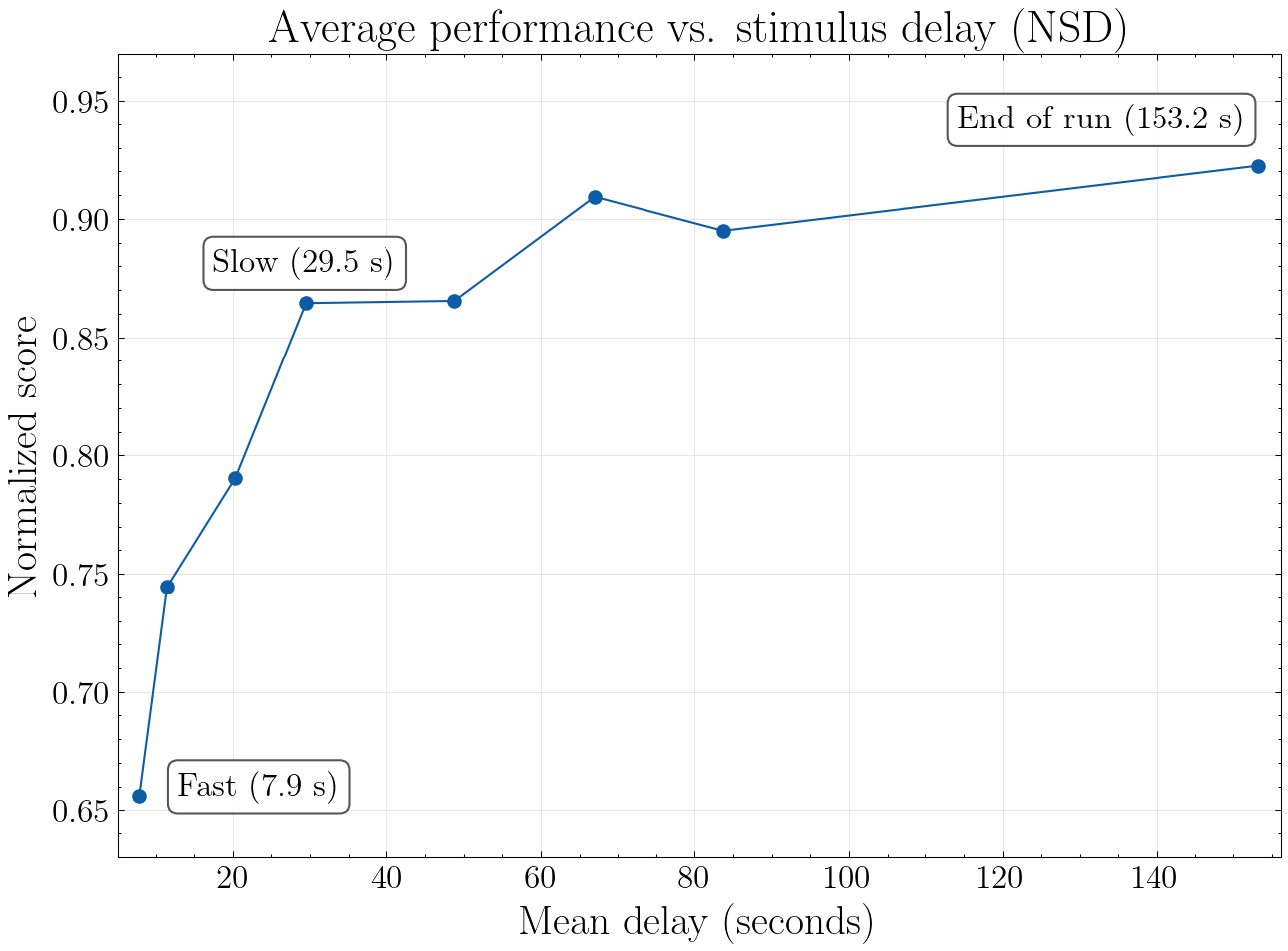}
  \caption{Performance (averaged across reconstruction/retrieval metrics listed in Table 1) across stimulus delays in NSD. Scores are min-max normalized per metric so that $0$ corresponds to using random COCO images as reconstructions and $1$ corresponds to offline NSD performance. Stimulus delay indicates the time elapsed after stimulus presentation before starting to analyze the neural response.}
  \label{fig:NSD-delay-vs-performance}
\end{figure}

We replicated the positive relationship between stimulus delay and performance (Figure \ref{fig:3t-delay-vs-performance}) using NSD subj01 (Figure \ref{fig:NSD-delay-vs-performance}).

\FloatBarrier

\subsection{Amount of Training Data vs. Performance in 3T and NSD}
\label{appendix-metrics-vs-training-data}
\begin{figure}
  \centering
  \includegraphics[width=0.6\textwidth]{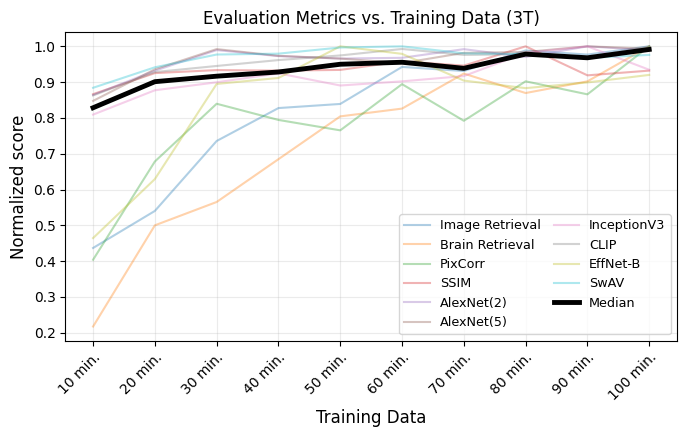}
  \caption{Evaluation metrics versus training data for the 3T subject. The model is fine-tuned with varying amounts of data from the 3T subject (up to two sessions) and evaluated using the ``fast real-time'' pipeline. Scores are normalized per metric so that $1$ corresponds to the score obtained from the largest amount of training data.}
  \label{fig:ablation-data-scaling-3t}
\end{figure}

\begin{figure}
  \centering
  \includegraphics[width=0.6\textwidth]{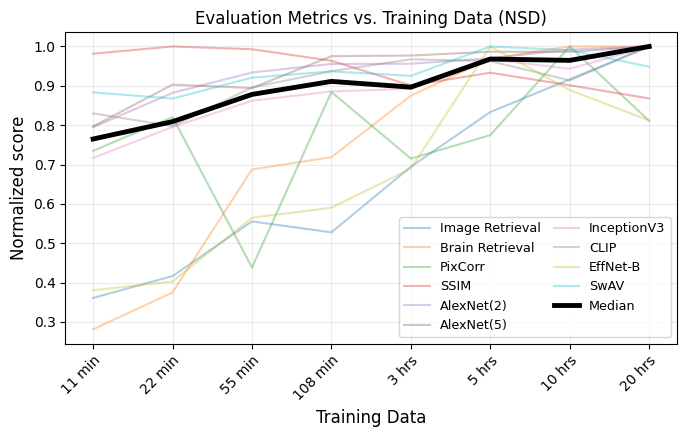}
  \caption{Evaluation metrics versus training data for NSD subj01. The model is fine-tuned with varying amounts of data from subj01 and evaluated using the ``fast real-time'' pipeline. Scores are normalized per metric so that $1$ corresponds to the score obtained from the largest amount of training data.}
  \label{fig:ablation-data-scaling-nsd}
\end{figure}

Figures \ref{fig:ablation-data-scaling-3t} and \ref{fig:ablation-data-scaling-nsd} report normalized reconstruction and retrieval metrics as a function of training data. In both cases, models are pretrained on 7 out of 8 NSD subjects. For Figure \ref{fig:ablation-data-scaling-3t}, the model is fine-tuned on varying amounts of data from the 3T participant. For Figure \ref{fig:ablation-data-scaling-nsd}, the model is fine-tuned on varying amounts of data from NSD subj01. Evaluation in both cases is performed on the same test set using the ``fast real-time'' pipeline. We observe that real-time performance improves with additional fine-tuning data in both datasets.

\FloatBarrier

\subsection{Detailed 3T Evaluations and Reconstructions}
\begin{figure*}[ht]
  \centering
  \includegraphics[width=\textwidth]{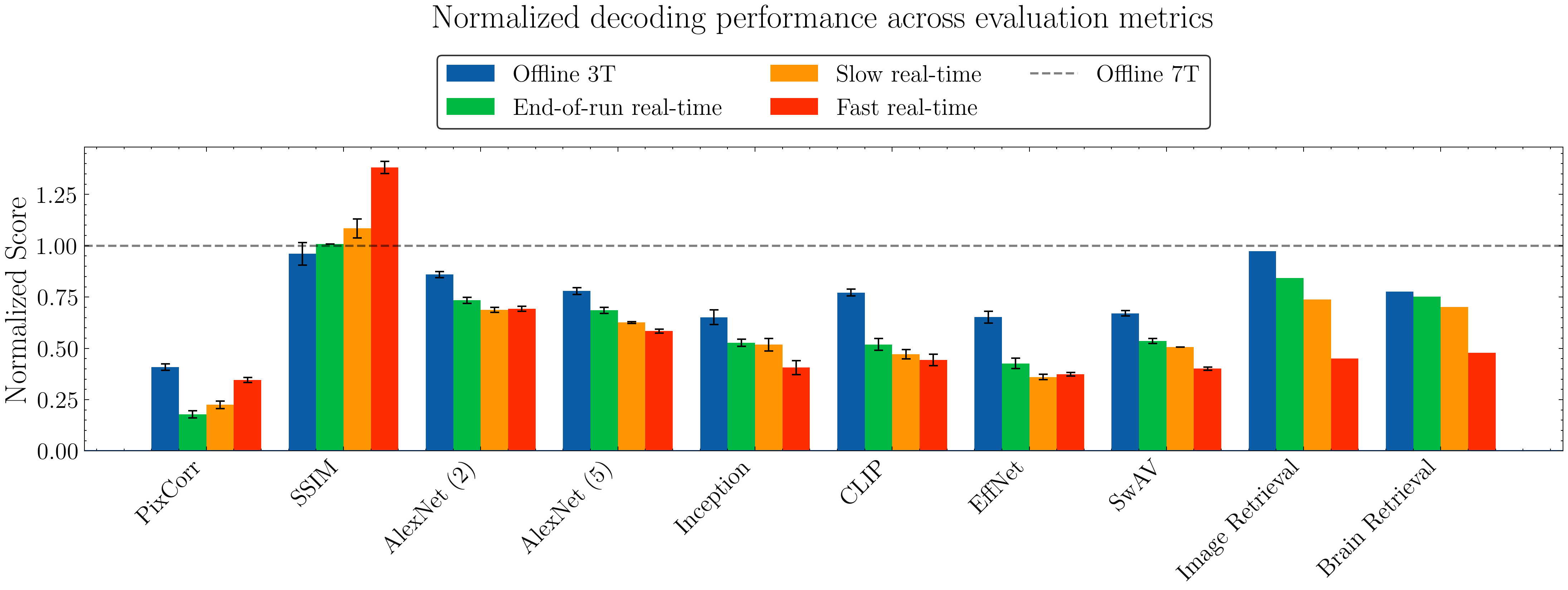}
  \caption{Normalized evaluation metrics for 3T offline and real-time pipelines. Scores are min-max normalized per metric so that $0$ corresponds to using random COCO images as reconstructions and $1$ corresponds to offline NSD performance. Bars show the mean and standard error across 5 random seeds.}
  \label{fig:barplot_eval_full}
\end{figure*}

\begin{figure}
  \centering
  \includegraphics[width=0.30\textwidth]{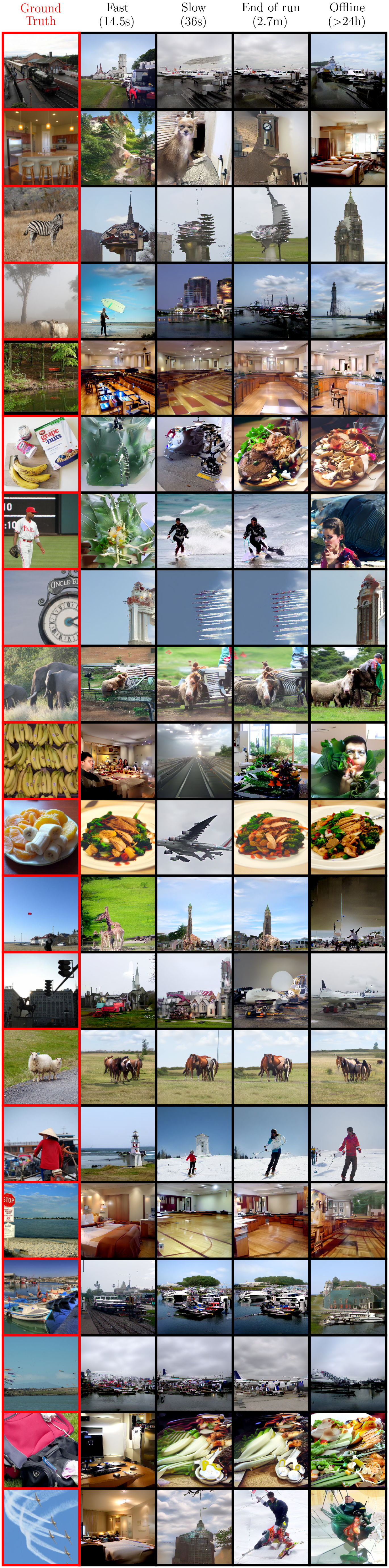}
  \caption{Randomly selected reconstructions for different configurations using 3T data.}
  \label{fig:large_grid_3T}
\end{figure}

Figure \ref{fig:barplot_eval_full} is a comprehensive version of Figure \ref{fig:barplot_eval_aggregated} (which aggregates metrics into low-level, high-level, and retrieval). Figure \ref{fig:large_grid_3T} contains additional, randomly chosen reconstructions as an extension of Figure \ref{fig:example_recons}.

\FloatBarrier

\FloatBarrier

\subsection{Experimenting with a Smaller Architecture} \label{appendix-arch-expt}
The CLIP space used in the experiments described above is very high dimensional ($\sim$426k dimensions). We considered the possibility that using a lower prediction dimensionality would be more appropriate given the limited amount of fine-tuning data and lower data quality (resulting from use of a 3T scanner and real-time compatible preprocessing). To address this, we modified the MindEye2 architecture to use the last layer of OpenCLIP ViT-H/14 \cite{radfordLearningTransferableVisual2021,dosovitskiyImageWorth16x162021}, which contains just 1024 dimensions. We then generated reconstructions using SDXL Turbo \cite{sauerAdversarialDiffusionDistillation2025}. We reasoned that, while using lower-dimensional CLIP embeddings might limit the ceiling of expressivity of the model, this approach would enforce a stronger naturalistic prior on the reconstructed images and thus make the reconstructions appear more clear and visually appealing. 

\begin{figure}
  \centering
  \includegraphics[width=0.30\textwidth]{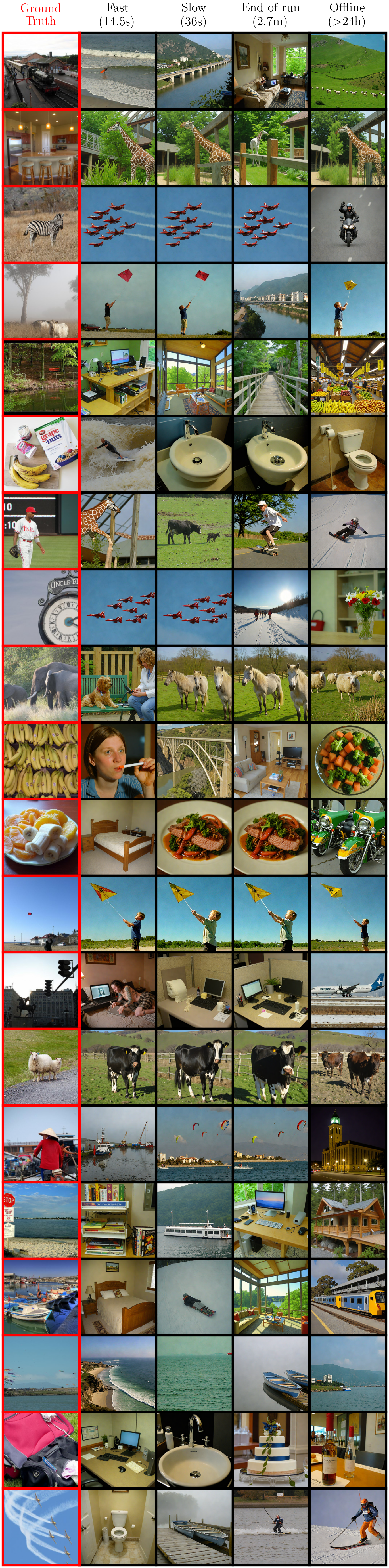}
  \caption{Randomly selected reconstructions for different configurations in the 3T subject, using SDXL Turbo.}
  \label{fig:large_grid_3T_sdxlturbo}
\end{figure}

While the generated reconstructions from this architecture (Figure \ref{fig:large_grid_3T_sdxlturbo}) indeed appeared more naturalistic and visually coherent as we intended, we found that, contrary to our intuition, the decoded contents were less faithful to the target images. This subjective assessment was supported by the quantitative metrics for reconstruction and retrieval (Table \ref{tab:recon_eval_sdxlturbo}). 

Finally, as a simple decoding baseline, we trained a small 9M parameter MLP (2048 hidden size) using MSE loss to predict CLIP embeddings (1664 dimension; pooled over the sequence dimension) from the training session, then evaluated on the held-out session with real-time processing. This achieves $\sim$6\% retrieval accuracy with a candidate pool size of 50 (1/50 = 2\%). This suggests that it is possible to reach above-chance real-time retrieval with a simple model, but that there is a substantial gain by using a more complex model (such as MindEye2).

\begin{table*}[ht]
    \centering
    \captionsetup{font=small}
    \setlength{\tabcolsep}{1pt}
    \small
    \resizebox{0.95\columnwidth}{!}{
    \begin{tabular}{lccccccccccc}
        \toprule
        Method & Latency & \multicolumn{4}{c}{Low-Level} & \multicolumn{4}{c}{High-Level} & \multicolumn{2}{c}{Retrieval}\\
        \cmidrule(lr){3-6} \cmidrule(lr){7-10} \cmidrule(l){11-12}
         & & PixCorr $\uparrow$ & SSIM $\uparrow$ & Alex(2) $\uparrow$ & Alex(5) $\uparrow$ & Incep $\uparrow$ & CLIP $\uparrow$  & Eff $\downarrow$ & SwAV $\downarrow$ & Image $\uparrow$ & Brain $\uparrow$\\
        \midrule
        Offline 3T (avg. 3 reps.) & $1$d & $0.082$ & $0.360$ & $67.3\%$ & $76.6\%$ & $64.4\%$ & $71.2\%$ & $0.908$ & $0.572$ & $14\%$ & $22\%$\\
        \midrule
        Offline 3T & $1$d & $0.080$ & $0.348$ & $61.4\%$ & $67.9\%$ & $59.4\%$ & $66.3\%$ & $0.920$ & $0.596$ & $10\%$ & $4\%$ \\
        End-of-run real-time & $2.7$m& $0.054$ & $0.352$ & $61.6\%$ & $68.5\%$ & $60.0\%$ & $61.6\%$ & $0.942$ & $0.604$ & $12\%$ & $2\%$ \\
        Slow real-time & $36$s& $0.062$ & $0.354$ & $60.8\%$ & $67.1\%$ & $57.5\%$ & $58.9\%$ & $0.938$ & $0.610$ & $6\%$ & $0\%$ \\
        Fast real-time & $14.5$s& $0.042$ & $0.346$ & $61.0\%$ & $61.5\%$ & $57.5\%$ & $57.8\%$ & $0.952$ & $0.632$ & $12\%$ & $4\%$ \\
        \bottomrule
    \end{tabular}}
    \caption{Latency and reconstruction/retrieval metrics for SDXL Turbo using 3T offline and real-time pipelines (reconstruction reported as the mean over 5 seeds; retrieval is deterministic). $\uparrow$ ($\downarrow$) means higher (lower) scores are better.}
    \label{tab:recon_eval_sdxlturbo}
\end{table*}

\FloatBarrier

\subsection{3T MRI Acquisition} \label{appendix-mri-acq}
We acquired two spin-echo field map volumes (TR = 8000 msec, TE = 66 msec) in opposite phase encoding directions for fieldmap correction. We collected two whole-brain T1-weighted (T1w) MPRAGE images (one from each fMRI scan session; TR = 2300 msec, TE = 2.98 msec, voxel size = 1 mm isotropic, flip angle = 9°, 176 slices, Generalized Autocalibrating Partially Parallel Acquisitions [GRAPPA] acceleration factor = 2) as well as two T2-weighted turbo spin-echo (TSE) images (one from each session; TR = 11390 msec, TE = 90 msec, voxel size = 0.44 × 0.44 × 1.5 mm, flip angle = 150°, 54 slices acquired perpendicular to the long axis of the hippocampus, distance factor = 20\%).

\subsection{Shifting Image Labels for Real-time} \label{tr-labels}
Prior to the start of the session, we assign an image label to each TR, shifted by $\sim$7.9 seconds post stimulus onset (for the ``fast'' real-time variation) to account for the BOLD response \cite{huettelFunctionalMagneticResonance2004}. As a result, prior to analysis, each TR is assigned either a label of a previously seen image or is simply left blank. To attempt image reconstruction or retrieval, the analysis waits until the TR with the assigned image label appears, using all data collected up to that point. Varying the stimulus delay (e.g., between fast, slow, and end-of-run variations) therefore involves simply shifting these TR labels. 

\clearpage
\glsaddallunused
\printnoidxglossaries
\clearpage
\section{fMRIPrep Preprocessing Details}
\label{fmriprep}
Results included in this manuscript come from preprocessing performed
using \emph{fMRIPrep} 24.0.1 (\citet{fmriprep1}; \citet{fmriprep2};
RRID:SCR\_016216), which is based on \emph{Nipype} 1.8.6
(\citet{nipype1}; \citet{nipype2}; RRID:SCR\_002502).

\begin{description}
\item[Preprocessing of B0 inhomogeneity mappings]
A total of 3 fieldmaps were found available within the input BIDS
structure for this particular subject. A \emph{B0}-nonuniformity map (or
\emph{fieldmap}) was estimated based on two (or more) echo-planar
imaging (EPI) references with \texttt{topup} (\citet{topup}; FSL None).
\item[Anatomical data preprocessing]
A total of 3 T1-weighted (T1w) images were found within the input BIDS
dataset. Each T1w image was corrected for intensity non-uniformity (INU)
with \texttt{N4BiasFieldCorrection} \citep{n4}, distributed with ANTs
2.5.1 \citep[RRID:SCR\_004757]{ants}. The T1w-reference was then
skull-stripped with a \emph{Nipype} implementation of the
\texttt{antsBrainExtraction.sh} workflow (from ANTs), using OASIS30ANTs
as target template. Brain tissue segmentation of cerebrospinal fluid
(CSF), white-matter (WM) and gray-matter (GM) was performed on the
brain-extracted T1w using \texttt{fast} \citep[FSL (version unknown),
RRID:SCR\_002823,][]{fsl_fast}. An anatomical T1w-reference map was
computed after registration of 3 T1w images (after INU-correction) using
\texttt{mri\_robust\_template} \citep[FreeSurfer 7.3.2,][]{fs_template}.
Volume-based spatial normalization to one standard space
(MNI152NLin2009cAsym) was performed through nonlinear registration with
\texttt{antsRegistration} (ANTs 2.5.1), using brain-extracted versions
of both T1w reference and the T1w template. The following template was selected for spatial normalization and accessed with
\emph{TemplateFlow} \citep[24.2.0,][]{templateflow}: \emph{ICBM 152
Nonlinear Asymmetrical template version 2009c}
{[}\citet{mni152nlin2009casym}, RRID:SCR\_008796; TemplateFlow ID:
MNI152NLin2009cAsym{]}.
\item[Functional data preprocessing]
For each of the 38 BOLD runs found per subject (across all tasks and
sessions), the following preprocessing was performed. First, a reference
volume was generated, using a custom methodology of \emph{fMRIPrep}, for
use in head motion correction. Head-motion parameters with respect to
the BOLD reference (transformation matrices, and six corresponding
rotation and translation parameters) are estimated before any
spatiotemporal filtering using \texttt{mcflirt} \citep[FSL
,][]{mcflirt}. The estimated \emph{fieldmap} was then aligned with
rigid-registration to the target EPI (echo-planar imaging) reference
run. The field coefficients were mapped on to the reference EPI using
the transform. The BOLD reference was then co-registered to the T1w
reference using \texttt{mri\_coreg} (FreeSurfer) followed by
\texttt{flirt} \citep[FSL ,][]{flirt} with the boundary-based
registration \citep{bbr} cost-function. Co-registration was configured
with six degrees of freedom. Several confounding time-series were
calculated based on the \emph{preprocessed BOLD}: framewise displacement
(FD), DVARS and three region-wise global signals. FD was computed using
two formulations following Power (absolute sum of relative motions,
\citet{power_fd_dvars}) and Jenkinson (relative root mean square
displacement between affines, \citet{mcflirt}). FD and DVARS are
calculated for each functional run, both using their implementations in
\emph{Nipype} \citep[following the definitions by][]{power_fd_dvars}.
The three global signals are extracted within the CSF, the WM, and the
whole-brain masks. Additionally, a set of physiological regressors were
extracted to allow for component-based noise correction
\citep[\emph{CompCor},][]{compcor}. Principal components are estimated
after high-pass filtering the \emph{preprocessed BOLD} time-series
(using a discrete cosine filter with 128s cut-off) for the two
\emph{CompCor} variants: temporal (tCompCor) and anatomical (aCompCor).
tCompCor components are then calculated from the top 2\% variable voxels
within the brain mask. For aCompCor, three probabilistic masks (CSF, WM
and combined CSF+WM) are generated in anatomical space. The
implementation differs from that of Behzadi et al.~in that instead of
eroding the masks by 2 pixels on BOLD space, a mask of pixels that
likely contain a volume fraction of GM is subtracted from the aCompCor
masks. This mask is obtained by thresholding the corresponding partial
volume map at 0.05, and it ensures components are not extracted from
voxels containing a minimal fraction of GM. Finally, these masks are
resampled into BOLD space and binarized by thresholding at 0.99 (as in
the original implementation). Components are also calculated separately
within the WM and CSF masks. For each CompCor decomposition, the
\emph{k} components with the largest singular values are retained, such
that the retained components' time series are sufficient to explain 50
percent of variance across the nuisance mask (CSF, WM, combined, or
temporal). The remaining components are dropped from consideration. The
head-motion estimates calculated in the correction step were also placed
within the corresponding confounds file. The confound time series
derived from head motion estimates and global signals were expanded with
the inclusion of temporal derivatives and quadratic terms for each
\citep{confounds_satterthwaite_2013}. Frames that exceeded a threshold
of 0.5 mm FD or 1.5 standardized DVARS were annotated as motion
outliers. Additional nuisance timeseries are calculated by means of
principal components analysis of the signal found within a thin band
(\emph{crown}) of voxels around the edge of the brain, as proposed by
\citep{patriat_improved_2017}. All resamplings can be performed with
\emph{a single interpolation step} by composing all the pertinent
transformations (i.e.~head-motion transform matrices, susceptibility
distortion correction when available, and co-registrations to anatomical
and output spaces). Gridded (volumetric) resamplings were performed
using \texttt{nitransforms}, configured with cubic B-spline
interpolation.
\item[Functional data preprocessing]
For each of the 38 BOLD runs found per subject (across all tasks and
sessions), the following preprocessing was performed. First, a reference
volume was generated, using a custom methodology of \emph{fMRIPrep}, for
use in head motion correction. Head-motion parameters with respect to
the BOLD reference (transformation matrices, and six corresponding
rotation and translation parameters) are estimated before any
spatiotemporal filtering using \texttt{mcflirt} \citep[FSL
,][]{mcflirt}. The BOLD reference was then co-registered to the T1w
reference using \texttt{mri\_coreg} (FreeSurfer) followed by
\texttt{flirt} \citep[FSL ,][]{flirt} with the boundary-based
registration \citep{bbr} cost-function. Co-registration was configured
with six degrees of freedom. Several confounding time-series were
calculated based on the \emph{preprocessed BOLD}: framewise displacement
(FD), DVARS and three region-wise global signals. FD was computed using
two formulations following Power (absolute sum of relative motions,
\citet{power_fd_dvars}) and Jenkinson (relative root mean square
displacement between affines, \citet{mcflirt}). FD and DVARS are
calculated for each functional run, both using their implementations in
\emph{Nipype} \citep[following the definitions by][]{power_fd_dvars}.
The three global signals are extracted within the CSF, the WM, and the
whole-brain masks. Additionally, a set of physiological regressors were
extracted to allow for component-based noise correction
\citep[\emph{CompCor},][]{compcor}. Principal components are estimated
after high-pass filtering the \emph{preprocessed BOLD} time-series
(using a discrete cosine filter with 128s cut-off) for the two
\emph{CompCor} variants: temporal (tCompCor) and anatomical (aCompCor).
tCompCor components are then calculated from the top 2\% variable voxels
within the brain mask. For aCompCor, three probabilistic masks (CSF, WM
and combined CSF+WM) are generated in anatomical space. The
implementation differs from that of Behzadi et al.~in that instead of
eroding the masks by 2 pixels on BOLD space, a mask of pixels that
likely contain a volume fraction of GM is subtracted from the aCompCor
masks. This mask is obtained by thresholding the corresponding partial
volume map at 0.05, and it ensures components are not extracted from
voxels containing a minimal fraction of GM. Finally, these masks are
resampled into BOLD space and binarized by thresholding at 0.99 (as in
the original implementation). Components are also calculated separately
within the WM and CSF masks. For each CompCor decomposition, the
\emph{k} components with the largest singular values are retained, such
that the retained components' time series are sufficient to explain 50
percent of variance across the nuisance mask (CSF, WM, combined, or
temporal). The remaining components are dropped from consideration. The
head-motion estimates calculated in the correction step were also placed
within the corresponding confounds file. The confound time series
derived from head motion estimates and global signals were expanded with
the inclusion of temporal derivatives and quadratic terms for each
\citep{confounds_satterthwaite_2013}. Frames that exceeded a threshold
of 0.5 mm FD or 1.5 standardized DVARS were annotated as motion
outliers. Additional nuisance timeseries are calculated by means of
principal components analysis of the signal found within a thin band
(\emph{crown}) of voxels around the edge of the brain, as proposed by
\citep{patriat_improved_2017}. All resamplings can be performed with
\emph{a single interpolation step} by composing all the pertinent
transformations (i.e.~head-motion transform matrices, susceptibility
distortion correction when available, and co-registrations to anatomical
and output spaces). Gridded (volumetric) resamplings were performed
using \texttt{nitransforms}, configured with cubic B-spline
interpolation.
\end{description}

Many internal operations of \emph{fMRIPrep} use \emph{Nilearn} 0.10.4
\citep[RRID:SCR\_001362]{nilearn}, mostly within the functional
processing workflow. For more details of the pipeline, see
\href{https://fmriprep.readthedocs.io/en/latest/workflows.html}{the
section corresponding to workflows in \emph{fMRIPrep}'s documentation}.

\subsection{Copyright Waiver}\label{copyright-waiver}

The above boilerplate text was automatically generated by fMRIPrep with the express intention that users should copy and paste this text into their manuscripts \emph{unchanged}. It is released under the \href{https://creativecommons.org/publicdomain/zero/1.0/}{CC0} license.

\end{bibunit}

\end{document}